\documentclass{article} 
\PassOptionsToPackage{usenames,dvipsnames,dvinames}{xcolor}
\usepackage{nips14submit_e,times}
\usepackage[bookmarks=false]{hyperref}
\providecommand{\doarxiv}{true}
\usepackage{times}
\usepackage{graphicx} 
\usepackage{subfigure} 
\usepackage{cleveref}
\crefname{section}{§}{§§}
\Crefname{section}{§}{§§}

\usepackage[numbers]{natbib}
\renewcommand{\baselinestretch}{0.99}

\usepackage{microtype}
\usepackage{wrapfig}
\usepackage[font=scriptsize]{caption}

\usepackage{epsfig}
\usepackage{float}
\usepackage{amsmath}
\usepackage{amssymb}
\usepackage{amsthm}
\usepackage{amsfonts}
\usepackage{dsfont}
\usepackage{rotating}
\usepackage{algorithm}
\usepackage{algorithmic}
\usepackage{multirow}
\usepackage{multicol}
\usepackage{blindtext}
\newtheorem{lemma}{Lemma}

\newtheorem{theorem}{Theorem}
\newtheorem{definition}{Definition}
\newtheorem{proposition}{Proposition}
\newtheorem{corollary}{Corollary}

\newcommand\independent{\protect\mathpalette{\protect\independenT}{\perp}}
\def\independenT#1#2{\mathrel{\rlap{$#1#2$}\mkern2mu{#1#2}}}

\usepackage{ifthen}
\newboolean{isdraft}
\setboolean{isdraft}{false} 
\ifthenelse{\boolean{isdraft}}{%
\newcommand{\myaddcomment}[3]{{\color{#1}{\ensuremath{\langle\!\!\langle}{\bf {#2} :} {#3}\ensuremath{\rangle\!\!\rangle}}}}
\newcommand{\tianyi}[1]{\myaddcomment{LimeGreen}{Tianyi}{#1}}
\newcommand{\JTT}[1]{\myaddcomment{LimeGreen}{Jeff\ensuremath{\rightarrow}Tianyi}{#1}}
\newcommand{\CTT}[1]{\myaddcomment{LimeGreen}{Carlos\ensuremath{\rightarrow}Tianyi}{#1}}
\newcommand{\jeff}[1]{\myaddcomment{blue}{Jeff}{#1}}
\newcommand{\TTJ}[1]{\myaddcomment{blue}{Tianyi\ensuremath{\rightarrow}Jeff}{#1}}
\newcommand{\CTJ}[1]{\myaddcomment{blue}{Carlos\ensuremath{\rightarrow}Jeff}{#1}}
\newcommand{\carlos}[1]{\myaddcomment{red}{Carlos}{#1}}
\newcommand{\JTC}[1]{\myaddcomment{red}{Jeff\ensuremath{\rightarrow}Carlos}{#1}}
\newcommand{\TTC}[1]{\myaddcomment{red}{Tianyi\ensuremath{\rightarrow}Carlos}{#1}}
}{
\newcommand{\tianyi}[1]{}
\newcommand{\JTT}[1]{}
\newcommand{\CTT}[1]{}
\newcommand{\jeff}[1]{}
\newcommand{\TTJ}[1]{}
\newcommand{\CTJ}[1]{}
\newcommand{\carlos}[1]{}
\newcommand{\JTC}[1]{}
\newcommand{\TTC}[1]{}
}

\providecommand{\doarxiv}{true}
\newboolean{isarxiv}
\setboolean{isarxiv}{\doarxiv} 
\ifthenelse{\boolean{isarxiv}}{%
\newcommand{\arxiv}[1]{{\color{gray}{#1}}}
\newcommand{\notarxiv}[1]{}
}{
\newcommand{\notarxiv}[1]{#1}
\newcommand{\arxiv}[1]{}
}
\newcommand{\arxivalt}[2]{\ifthenelse{\boolean{isarxiv}}{#1}{#2}}

\title{Divide-and-Conquer Learning by Anchoring a Conical Hull}

\author{
Tianyi Zhou$^\dagger$, Jeff Bilmes$^\ddagger$, Carlos Guestrin$^\dagger$\\
$^\dagger$Computer Science \& Engineering, $^\ddagger$Electrical Engineering\\
University of Washington, Seattle\\
\texttt{\{tianyizh, bilmes, guestrin\}@uw.edu} \\
}

\nipsfinalcopy 

\newcommand{\cone}{\operatorname{cone}}
\newcommand{\cover}{\operatorname{cover}}
\newcommand{\MCH}{\operatorname{MCH}}
\newcommand{\DCA}{\operatorname{DCA}}

\begin{document}

\maketitle

\begin{abstract}

We reduce a broad class of machine learning problems, usually addressed by EM or sampling, to the problem of finding the $k$ extremal rays spanning the conical hull of a data point set. These $k$ ``anchors'' lead to a global solution and a more interpretable model that can even outperform EM and sampling on generalization error. To find the $k$ anchors, we propose a novel divide-and-conquer learning scheme ``DCA'' that distributes the problem to $\mathcal O(k\log k)$ same-type sub-problems on different low-D random hyperplanes, each can be solved by any solver. For the 2D sub-problem, we present a non-iterative solver that only needs to compute an array of cosine values and its max/min entries. DCA also provides a faster subroutine for other methods to check whether a point is covered in a conical hull, which improves algorithm design in multiple dimensions and brings significant speedup to learning. We apply our method to GMM, HMM, LDA, NMF and subspace clustering, then show its competitive performance and scalability over other methods on rich datasets.
\end{abstract}

\section{Introduction}

Expectation-maximization (EM)~\cite{DempsterEM}, sampling methods~\cite{GibbsSampling}, and matrix factorization~\cite{NMF,PMF} are three algorithms commonly used to produce maximum likelihood (or maximum a posteriori (MAP)) estimates of models with latent variables/factors, and thus are used in a wide range of applications such as clustering, topic modeling, collaborative filtering, structured prediction, feature engineering, and time series analysis. However, their learning procedures rely on alternating optimization/updates between parameters and latent variables, which suffer from local optima. Hence, their quality greatly depends on initialization and on using a large number of iterations for proper convergence~\cite{MMlocal}. 

The method of moments~\cite{MMoments,BelkinMoment,KalaiMoment}, in contrast, solves $m$ equations by relating the first $m$ moments of observation $x\in\mathbb R^p$ to the $m$ model parameters, and thus yields a consistent estimator with a global solution. In practice, however, sample moments usually suffer from unbearably large variance, which easily leads to the failure of final estimation, especially when $m$ or $p$ is large. Although recent spectral methods \cite{ChangSpectral,VempalaSectral,SpectralHMM,SpectralLDA} reduces $m$ to $2$ or $3$ when estimating $\mathcal O(p)\gg m$ parameters  \cite{SpectralMM} by relating the eigenspace of lower-order moments to parameters in a matrix form up to column scale, the variance of sample moments is still sensitive to large $p$ or data noise, which may result in poor estimation. Moreover, although spectral method using SVDs or tensor decomposition evidently simplifies learning, the computation can still be expensive for big data. In addition, recovering a parameter matrix with uncertain column scale might not be feasible for some applications.

In this paper, we reduce the learning in a rich class of models (e.g., matrix factorization and latent variable model) to finding the extreme rays of a conical hull from a finite set of real data points. This is obtained by applying a general \emph{separability assumption} to either the data matrix in matrix factorization or the $2^{nd}$/$3^{rd}$ order moments in latent variable models. Separability posits that a set of $n$ points, as rows of matrix $X$, can be represented by $X=FX_A$, where the rows(bases) in $X_A$ are a subset $A \subset V=[n]$ of rows in $X$, which are called ``anchors'' and are interesting to various models when $|A|=k\ll n$. This property was introduced in \cite{UniqueNMF} to establish the uniqueness of non-negative matrix factorization (NMF) under simplex constraints, and was later~\cite{XRAY,SPA} extended to non-negative constraints. We generalize it further to the model $X=FY_A$ for two (possibly distinct) finite sets of points $X$ and $Y$, and build a new theory for the identifiability of $A$. This generalization enables us to apply it to more general models (ref. Table \ref{table:models}) besides NMF. More interestingly, it leads to a learning method with much higher tolerance to the variance of sample moments or data noise, a unique global solution, and a more interpretable model.


\notarxiv{ 
\begin{wrapfigure}{r}{0.4\linewidth}\vspace{-1mm}
\begin{center}\vspace{-8mm}
 \includegraphics[width=1\linewidth]{cone.pdf}
\end{center}\vspace{-3mm}
   \caption{\scriptsize{Geometry for general minimum conical hull problem and divide-and-conquer anchoring (DCA).}}
\label{fig:cone}
\vspace{-3mm}
\end{wrapfigure}
}

Another primary contribution of this paper is a distributed learning scheme, ``divide-and-conquer anchoring (DCA)'', for finding an anchor set $A$ such that $X=FY_A$ by solving same-type sub-problems on merely $\mathcal O(k\log k)$ random drawn low-dimensional (low-D) hyperplanes. Each sub-problem is of the form of $(X\Phi)=F\cdot (Y\Phi)$ with random projection matrix $\Phi$, and can easily be handled by most solvers due to the low dimension. This is based on the observation that the geometry of the original conical hull is partially preserved after a random projection. We analyze the probability of success for each sub-problem to recover part of $A$, and then study the number of sub-problems for recovering the whole $A$ with high probability (\emph{w.h.p.}). In particular, we propose an ultrafast non-iterative solver for sub-problems on the 2D plane, which requires computing an array of cosines and its max/min values, and thus results in learning algorithms with speedups of tens to hundreds of times. DCA improves multiple aspects of algorithm design since: 1) its idea of divide-and-conquer randomization gives rise to distributed learning that can reduce the original problem to multiple extremely low-D sub-problems that are much easier and faster to solve, and 2) it provides a fast subroutine checking if a point is covered by a conical hull, which can be embedded into other solvers.

We apply both the conical hull anchoring model and DCA to five learning models: Gaussian mixture models (GMM)~\cite{MMbook}, hidden Markov models (HMM)~\cite{HMM}, latent Dirichlet allocation (LDA)~\cite{LDA}, NMF~\cite{NMF}, and subspace clustering (SC)~\cite{SSC}. The resulting models and algorithms show significant improvement in efficiency. On generalization performance, they consistently outperform spectral methods and matrix factorization, and are comparable to or even better than EM and sampling.

In the following, we will first generalize the separability assumption and minimum conical hull problem risen from NMF in \S\cref{sec:2}, and then show how to reduce more general learning models to a (general) minimum conical hull problem in \S\cref{sec:3}. \S\cref{sec:4} presents a divide-and-conquer learning scheme that can quickly locate the anchors of the conical hull by solving the same problem in multiple extremely low-D spaces. Comprehensive experiments and comparison can be found in \S\cref{sec:5}.

\vspace{-4mm}
\section{General Separability Assumption and Minimum Conical Hull Problem}\label{sec:2}
\vspace{-3mm}
The original separability property \cite{UniqueNMF} is defined on the convex hull of a set of data points, namely that each point can be represented as a convex combination of certain subsets of vertices that define the convex hull.
Later works on separable NMF \cite{XRAY,SPA} extend it to the conical hull case, which replaced convex with conical combinations. Given the definition of (convex) cone and conical hull, the separability assumption can be defined both geometrically and algebraically.
\begin{definition}[\bf {Cone \& conical hull}]
A (convex) cone is a non-empty convex set that is closed with respect to conical combinations of its elements. In particular, $\cone(R)$ can be defined by its $k$ generators (or rays) $R=\{r_i\}_{i=1}^k$ such that 
\begin{equation}\label{equ:cone}
\cone(R)=\{\sum\nolimits_{i=1}^k\alpha_ir_i\mid r_i\in R,\alpha_i\in\mathbb R_+ \, \forall i\}.
\end{equation}
\end{definition}

\arxivalt{
In the following, let $X$ be a matrix with $n$ columns and indexed by the integers $[n]$,
and $X_A$ be a subset of columns for $A \subseteq [n]$.
\begin{definition}[\bf Separability assumption] 
\label{def:2}
All the data points in $X$ are covered in a finitely generated and pointed cone whose generators are a subset $A\subseteq[n]$ of data points.
That is, 
\begin{equation}\label{equ:sepag}
\exists A \subseteq [n] \text{ s.t. } \forall i\in[n], X_i\in \cone\left(X_A\right), X_A=\{x_i\}_{i\in A}.
\end{equation}
An equivalent algebraic form is
$
X=FX_A, \Pi F=\left[
                          \begin{array}{c}
                            I_k \\
                            F' \\
                          \end{array}
                        \right]
$,
where $|A|=k$, $I_k$ is a $k$-by-$k$ identity matrix, $F'\in\mathbb R_+^{(n-k)\times k}$, and $\Pi$ is a row permutation matrix.
\end{definition}
The above algebraic form gives rise to an NMF model $X=FX_A$. The cone $\cone(X_A)$ is \emph{finitely generated} because all the elements in $X$ are conical combinations of a finite set $X_A$. It is also \emph{pointed} because its non-negativity does not allow it containing both $x$ and $-x$. According to basic rules \cite{ConicalHull}, \emph{a finitely generated and pointed cone $C$ possesses a finite and unique set of extreme rays $R$, and $C = \cone(R)$ is the conical hull generated by these extreme rays $R$}. When only one point in $X$ is on each extreme ray, the separability assumption guarantees the uniqueness of NMF solution $X_A$ and $F$. Moreover, when
$X$ consists of real data points, they are often in practice interpretable
since they constitute the ``essential'' set that uses actual data to express itself rather than artificial basis/factors.\JTT{I changed the language
of this to reflect what you mean, but it would be useful either to add
refs for where this is the case, or forward reference to below in the paper
where this is indeed the case.}\TTJ{This is an empirical observation, hard to find reference, the only thing I can remember is Arora mentioned in his paper that David Blei talked to him that ``anchor word'' is very common in topic modeling. Other people just claim that basis selected from real data is more explanable.}
\looseness-1



Based on separability assumption in Definition \ref{def:2}, we define the minimum conical hull problem, and offer
several formulations for how it can be solved, including a novel
formulation that utilizes submodular optimization. Each of these, in
later sections of the paper, will be compared in various ways against
the baseline methods established in this section.

\begin{definition}[\bf{Minimum Conical Hull Problem}]
\label{def:min_con_hull_problem}
Given a $X$ having an index set $V=[n]$ of its rows, minimum conical hull problem finds the subset of rows that define the same cone as all the rows. That is,
find $A\in 2^V$ that solves:
\begin{equation}
\min_{A \subset V} |A|, \text{ s.t., } \cone(X_A) = \cone(X).
\end{equation}
where $\cone(X_A)$ is the cone induced by the rows $A$ of $X$.
\end{definition}

We note that without the separability assumption, there is no non-trivial set $A \subset V$ that is feasible. 
As mentioned above, there is a weakly polynomial time algorithm that
in the non-negative case 
solves the problem via backward removal.
\begin{definition}
  A nonnegative matrix $X_A$ is \emph{simplicial} if no row in $X_A$
  can be represented in the convex hull of the remaining rows in
  $X_A$.
\end{definition}

\begin{lemma}[Lemma 5.3 from \cite{AGKM}]
If a nonnegative matrix $X$ has a separable factorization $WX_A$ of inner-dimension at most $k$ then there is one in which $X_A$ is simplicial. 
\end{lemma}
In the proof of this lemma, Arora proposes a procedure that starting
at $A=V$, one-by-one, removes rows violating the simpliciality of
$X_A$.  This is using linear programming, where for each row in $X_A$
we test if it can be represented by convex combination of remaining
rows. The rows that cannot be represented by other rows are selected as the ``loners'' (anchors).
\begin{theorem}[Theorem 5.4 in \cite{AGKM}]
There is an algorithm that runs in time \arxiv{[weakly]} polynomial in $n$, $m$, and $k$ and given a matrix $X$ outputs a separable factorization with inner dimension at most $k$ (if one exists).
\end{theorem}

The simplicial $X_A$ obtained by the algorithm is the
solution of our minimum \emph{conical} hull problem.
\begin{theorem}
A simplicial $X_A$ satisfying $X=FX_A, F_{i,j}\geq 0~~\forall \{i,j\}$ is the solution of~\eqref{def:min_con_hull_problem}. Such $X_A$ can be achieved by Arora's algorithm that runs in time polynomial in $n$, $m$, and $k$.
\end{theorem}
\begin{proof}
It is easy to verify that $X_A$ obtained by Arora's exact algorithm \cite{AGKM} is a feasible solution fulfilling the constraint in (\ref{equ:mch1}). However, suppose (for contradiction) $X_A$ does not define the minimum conical hull, i.e., there exists a simplicial $X_B$ with $|B|<|A|$ and $X=F^{'}X_B$. So we have
\begin{equation}
X_A=CX_B, X_B=DX_A\rightarrow X_A=(CD)X_A.
\end{equation}
If $U=CD$, for arbitrary row $X_A^j$ in $X_A$, we have
\begin{equation}
X_A^j=U^j_{\backslash j}X_{A\backslash j} + U^j_jX_A^j.
\end{equation}
We cannot make $U^j_j=1\forall j$, because in this case $CD=I_{|A|}$, which leads to contradiction with the fact that $CD$ has inner dimension of $|B|$. Therefore, we can always find at least one $j$ such that $U^j_{\backslash j}$ is not all-zero vector and
\begin{equation}
X_A^j = \frac{U^j_{\backslash j}X_{A\backslash j}}{1-U^j_j},
\end{equation}
which violates the constraint that $X_A$ is simplicial and thus causes contradiction. So we cannot find a simplicial $X_B$ with $|B|<|A|$ and $X=W^{'}X_B$, which implies $A$ is the solution of minimum conical hull problem (\ref{equ:mch1}).
\end{proof}

Lastly, we see how Equation~\eqref{def:min_con_hull_problem} can be
seen as a submodular cover problem \cite{wolsey1982analysis}.
Define $\cover(X_A)$ to be the size of the largest set $B \subseteq V$ of rows of $X$ that lies within $\cone(X_A)$. That is, 
\begin{align}
  \cover(X_A)
  = |\{ v \in V : \exists f \in \mathbb R^{|A|}_+  \text{ with } X_v = f^\top X_A \} |.
\end{align}
\begin{definition}
For a nonnegative matrix $X$ with a ground set $V$ of all its row indexes, the minimum conical hull problem is to solve:
\begin{equation}
\label{equ:mch1}
\min_{A \subset V} |A|, s.t., \cover(A) = \cover(V).
\end{equation}
\label{def:cover_variant}
\end{definition}
It can be verified that the constraint in
Definition~\ref{def:cover_variant} equals to the constraint to $F$ in
Definition~\ref{def:2}, and that $f(A) = \cover(X_{A \cup I})$,
where $X_I$ is any set of linearly independent row vectors,
is submodular \cite{Fujishige}. Therefore, we can apply efficient greedy algorithm for submodular cover to the minimum conical hull problem with approximation guarantee. \looseness-1
}{See~\cite{nips2014supplemental} for the original separability assumption, equivalence between separable NMF and minimum conical hull problem, which is defined as a submodular set cover problem.}

\vspace{-2mm}
\subsection{General Separability Assumption and General Minimum Conical Hull Problem}

By generalizing the separability assumption, we obtain a general minimum conical hull problem that can reduce more general learning models besides NMF, e.g., latent variable models and matrix factorization, to finding a set of ``anchors'' on the extreme rays of a conical hull. \arxiv{At first, we only need $F$ to be non-negative, and do not require $X_A$ or $X$ to be non-negative. Because a pointed cone $cone(X_A)$ after rotation is still a pointed cone with unchanged anchor set $A$. Secondly, to ensure $cone(X_A)$ to be finitely generated, we only need the points in $X_A$ to be selected from a finite set, which could be another set $Y$ and is not necessary to be $X$ itself. Lastly, extra constraint can be imposed to the coefficients $F'$ to encourage particular structures among data points.}

\begin{definition}[\bf{General separability assumption}]
\label{def:3}
All the $n$ data points(rows) in $X$ are covered in a finitely generated and pointed cone (i.e., if $x\in\cone(Y_A)$ then $-x\not\in\cone(Y_A)$) whose generators form a subset $A\subseteq[m]$ of data points in $Y$ such that $\nexists i\neq j, Y_{A_i}=a\cdot Y_{A_j}$. Geometrically, it says
\begin{equation}\label{equ:sepa}
\forall i\in[n], X_i\in cone\left(Y_A\right), Y_A=\{y_i\}_{i\in A}.
\end{equation}
An equivalent algebraic form is
$
X=FY_A, \Pi F=\left[
                          \begin{array}{c}
                            I_k \\
                            F' \\
                          \end{array}
                        \right]
$,
where $|A|=k$, $I_k$ is a $k$-by-$k$ identity matrix, $F'\in S\subseteq\mathbb R_+^{(n-k)\times k}$, and $\Pi$ is a row permutation matrix.
\end{definition}

When $X=Y$ and $S=\mathbb R_+^{(n-k)\times k}$, it degenerates to the original separability assumption given in~\cite{nips2014supplemental}. We generalize the minimum conical hull problem from~\cite{nips2014supplemental}. Under the general separability assumption, it aims to find the anchor set $A$ from the points in $Y$ rather than $X$. 
\begin{definition}[\bf{General Minimum Conical Hull Problem}]
\label{def:general_min_con_hull_problem}
Given a finite set of points $X$ and a set $Y$ having an index set $V=[m]$ of its rows, the general minimum conical hull problem finds the subset of rows in $Y$ that define a super-cone for all the rows in $X$. That is,
find $A\in 2^V$ that solves:
\begin{equation}
\min_{A \subset V} |A|, \text{ s.t., } \cone(Y_A)\supseteq \cone(X).
\end{equation}
where $\cone(Y_A)$ is the cone induced by the rows $A$ of $Y$.
\end{definition}

When $X=Y$, it degenerates to the original minimum conical hull problem defined in~\cite{nips2014supplemental}. A critical question of the general one is whether/when the solution $A$ is unique. When $X=Y$ and $X=FX_A$, \arxiv{which is the case for matrix factorization and latent variable model with nonzero off-diagonal entries in $D$ in \S\cref{sec:32},} by following the analysis of the separability assumption in \cite{nips2014supplemental},
we can prove that $A$ is unique and identifiable given $X$. However, when $X\neq Y$ and $X=FY_A$, it is clear that there could be multiple legal choices of $A$ (e.g., there could be multiple layers of conical hulls containing a set of points $X$ in the center). Fortunately, when the rows of $Y$ are rank-one matrices after vectorization (concatenating all columns to a long vector), which is the common case in most latent variable models\arxiv{ with diagonal $D$} in \S\cref{sec:32}, $A$ can be uniquely determined if the number of rows in $X$ exceeds $2$. 

\begin{lemma}[\bf{Identifiability}]
\label{lemma:iden}
If $X=FY_A$ with the additional structure $Y_s={\rm vec}(O_i^s\otimes O_j^s)$ where $O_i$ is a $p_i\times k$ matrix and $O_i^s$ is its $s^{th}$ column, under the general separability assumption in Definition \ref{def:3}, two (non-identical) rows in $X$ are sufficient to exactly recover the unique $A$, $O_i$ and $O_j$. 
\end{lemma}
\arxivalt {
\begin{proof}
Two non-identical rows of $X$ can be represented as two sum mixtures of rank-one matrices in $Y_A$ with weights $a\neq b$, i.e.,
\begin{equation}\label{equ: X1X2}
\begin{array}{ll}
& X_1 = {\rm vec}(\sum_{s=1}^ka^{(s)}O_i^s\otimes O_j^s)={\rm vec}(O_i{\rm Diag}(a)O_j^T),\\
& X_2 = {\rm vec}(\sum_{s=1}^kb^{(s)}O_i^s\otimes O_j^s)={\rm vec}(O_i{\rm Diag}(b)O_j^T).
\end{array}
\end{equation}
In the following proof, we temporarily use $X_1$ and $X_2$ to denote themselves before vectorization. Given SVD of $X_1$ as $X_1=U_i\Sigma U_j^T$, $O_i$ and $O_j$ can be represented in the following forms.
\begin{equation}\label{equ:OiOj}
O_i=U_i\Sigma^{1/2}V{\rm Diag}(a^{-1/2}), O_j=U_j\Sigma^{1/2}V^{-T}{\rm Diag}(a^{1/2}).
\end{equation}
Note the decomposition of $X_1$ in (\ref{equ: X1X2}) will stay the same when we change $V$ in (\ref{equ:OiOj}) to any other non-singular matrix. Thus a single row in $X$ cannot uniquely identify $O_i$ and $O_j$. By further given $X_2$, we have
\begin{equation}\label{equ:2svd}
\begin{array}{ll}
\Sigma^{-1/2}U_i^TX_2U_j\Sigma^{-1/2}=\Sigma^{-1/2}U_i^TO_i{\rm Diag}(b)O_j^TU_j\Sigma^{-1/2}=V{\rm Diag}(b./a)V^{-1}.
\end{array}
\end{equation}
The second equality is obtained by applying (\ref{equ:OiOj}). Let the (unique) eigendecomposition of $V{\rm Diag}(b./a)V^{-1}$ to be $W\Lambda W^T$, we have $V=W{\rm Diag}(c)$. Substituting it into (\ref{equ:OiOj}) yields
\begin{equation}\label{equ:OiOj2}
O_i=U_i\Sigma^{1/2}W{\rm Diag}(c\cdot/\sqrt{a}), O_j=U_j\Sigma^{1/2}W^{-T}{\rm Diag}(\sqrt{a}\cdot/c).
\end{equation}
Therefore, we can determine each element $s\in A$ by checking if $\exists t, Y_s={\rm vec}(O_i^s\otimes O_j^s)={\rm vec}\left[(U_i\Sigma^{1/2}W)^{(t)}\otimes (U_j\Sigma^{1/2}W^{-T})^{(t)}\right]$ for each $s\in[n]$ in the ground set. Since all the quantities in the rank-one matrix on the right hand side is uniquely fixed for each $t\in[k]$, each $s$ selected from the ground set is unique.
\end{proof}

In applications such as latent variable model, as stated in \S\cref{sec:32}, the set of $n$ rank-one matrices in $Y$ is generated by computing $x_i\otimes x_j$ for $n$ data points or observations, where $x_i$ is the $i^{th}$ group of features. The general minimum conical hull problem then equals to selecting $k$ data points. Since in this case the columns of $O_i$ and $O_j$ are respectively estimated as $x_i$ and $x_j$ of the $k$ selected data points, in order to further guarantee the uniqueness of the $k$ selected data points rather than only the $k$ out products $x_i\otimes x_j$, we need to avoid the case when $x_i$ and $x_j$ in two different data points generate the same $x_i\otimes x_j$. According to (\ref{equ:OiOj2}), we have to assume $\nexists\{s\in A,t\neq s,\delta\neq 0\}$ satisfying $\{x_i\}_t=\delta \{x_i\}_s$ and $\{x_j\}_t=\{x_j\}_s/\delta$, where $\{x_i\}_s$ denotes feature $x_i$ for $s^{th}$ data point. Note this assumption is much weaker than limiting $\{x\}_t\neq\delta\{x\}_s$ in original separable NMF.

In above proof, we need $a\neq b$ to ensure the identifiability of $A$. In practice, this inequality can be achieved by drawing random vector $\eta$ from a continuous distribution, as we did in (\ref{equ: Dmatrix}) within \S\cref{sec:3}. Then the inequality holds with probability $1$.}{See~\cite{nips2014supplemental} for proof and additional uniqueness conditions when applied to latent variable models.} 

\vspace{-2mm}
\section{Minimum Conical Hull Problem for General Learning Models}\label{sec:3}

\begin{table}[htp]
\vspace{-3mm}
\tiny
\caption{\scriptsize{Summary of reducing NMF, SC, GMM, HMM and LDA to a conical hull anchoring model $X=FY_A$ in~\S\cref{sec:3}, and their learning algorithms achieved by $A=\DCA(X,Y,k,\mathbb M)$ in Algorithm \ref{A:DCA} . Minimal conical hull $A=\MCH(X,Y)$ is defined in Definition \ref{def:mch}. ${\rm vec}(\cdot)$ denotes the vectorization of a matrix. For GMM and HMM, $X_i\in\mathbb R^{n\times p_i}$ is the data matrix for view $i$ (i.e., a subset of features) and the $i^{th}$ observation of all triples of sequential observations, respectively. $X_{t,i}$ is the $t^{th}$ row of $X_i$ and associates with point/triple $t$. $\eta_t$ is a vector uniformly drawn from the unit sphere. More details are given in \cite{nips2014supplemental}.}}\vspace{-4mm}
\begin{center}
\begin{tabular}{l||l|l|l}
\hline
Model &$X$ in conical hull problem &$Y$ in conical hull problem &$k$ in conical hull problem\\
\hline
NMF &data matrix $X\in\mathbb R_{+}^{n\times p}$ &$Y:=X$	&\# of factors	\\
SC &data matrix $X\in\mathbb R^{n\times p}$ &$Y:=X$	&\# of basis from all clusters\\
GMM &$[{\rm vec}[X_1^TX_2]; {\rm vec}[X_1^T{\rm Diag}(X_3\eta_t)X_2]_{t\in[q]}]/n$ &$[{\rm vec}(X_{t,1}\otimes X_{t,2})]_{t\in[n]}$	&\# of components/clusters\\
HMM &$[{\rm vec}[X_2^TX_3]; {\rm vec}[X_2^T{\rm Diag}(X_1\eta_t)X_3]_{t\in[q]}]/n$ &$[{\rm vec}(X_{t,2}\otimes X_{t,3})]_{t\in[n]}$	&\# of hidden states\\
LDA &word-word co-occurrence matrix $X\in\mathbb R_{+}^{p\times p}$ &$Y:=X$	&\# of topics\\
\hline
\hline
Algo &Each sub-problem in DCA &Post-processing after $A:=\bigcup_i\tilde A^i$ &Interpretation of anchors indexed by $A$\\
\hline
NMF &$\tilde A=\MCH(X\Phi, X\Phi)$, can be solved by (\ref{equ:2Danchor})	&solving $F$ in $X=FX_A$ &basis $X_A$ are real data points\\
SC  &$\tilde A=$anchors of clusters achieved by \emph{meanshift}($\widehat{(X\Phi)\varphi}$) &clustering anchors $X_A$ &cluster $i$ is a cone $cone(X_{A_i})$\\
GMM &$\tilde A=\MCH(X\Phi, Y\Phi)$, can be solved by (\ref{equ:2Danchor}) &N/A	&centers $[X_{A, i}]_{i\in[3]}$ from real data \\
HMM &$\tilde A=\MCH(X\Phi, Y\Phi)$, can be solved by (\ref{equ:2Danchor}) &solving $T$ in $OT=X_{A,3}$	&emission matrix $O=X_{A,2}$\\
LDA &$\tilde A=\MCH(X\Phi, X\Phi)$, can be solved by (\ref{equ:2Danchor}) &col-normalize $\{F:X=FX_A\}$ 	&anchor word for topic $i$ (topic prob. $F_i$)\\
\hline
\end{tabular}\label{table:models}
\end{center}\vspace{-2mm}
\vspace{-1mm}
\end{table}
\normalsize

In this section, we discuss how to reduce the learning of general models such as matrix factorization and latent variable model to the (general) minimum conical hull problem. Five examples are given in Table \ref{table:models} to show how this general technique can be applied to specific models.

\subsection{Matrix Factorization}

Besides NMF, we consider more general matrix factorization (MF) models that can operate on negative features and specify a complicated structure of $F$. The MF $X=FW$ is a deterministic latent variable model where $F$ and $W$ are deterministic latent factors. By assigning a likelihood $p(X_{i,j}|F_i, (W^T)_j)$ and priors $p(F)$ and $p(W)$, its optimization model can be derived from maximum likelihood or MAP estimate. The resulting object is usually a loss function $\ell(\cdot)$ of $X-FW$ plus regularization terms for $F$ and $W$, i.e., $\min \ell(X,FW)+R_F(F)+R_W(W)$.

Similar to separable NMF, minimizing the objective of general MF can be reduced to a minimum conical hull problem that selects the subset $A$ with $X=FX_A$. In this setting, $R_W(W)=\sum_{i=1}^kg(W_i)$ where $g(w)=0$ if $w = X_i$ for some $i$ and $g(w)=\infty$ otherwise. This is equivalent to applying a prior $p(W_i)$ with finite support set on the rows of $X$ to each row of $W$. In addition, the regularization of $F$ can be transformed to geometric constraints between points in $X$ and in $X_A$. Since $F_{i,j}$ is the conical combination weight of $X_{A_j}$ in recovering $X_i$, a large $F_{i,j}$ intuitively indicates a small angle between $X_{A_j}$ and $X_i$, and vice verse. For example, the sparse and graph Laplacian prior for rows of $F$ in subspace clustering can be reduced to ``cone clustering'' finding $A$.

\arxivalt {

\subsection{Example: Subspace Clustering}\label{sec:SC}

Subspace clustering (SC) assumes that the data points in each cluster exist in a low-Dimensional subspace. The subspaces defining different clusters are assumed to be distinguishable, e.g., with large principle angles between each other~\cite{RSC}. SC outperforms traditional clustering methods in various tasks such as motion segmentation. Most existing subspace clustering methods~\cite{SSC,RSC}, relies on spectral clustering to the sparse representations of data points, which are obtained by finding a sparse $C$ with ${\rm diag}(C)=0$ in model $X=CX$. This usually requires a series of time costly lasso-type optimization. Then (spectral) clustering to rows of $C$ guarantees to provide clustering labels which can lead to reliable estimation of the $k$ subspaces~\cite{RSC}.

Under the general separability assumption, SC model can be reduced to general minimum conical hull problem $X=FX_A$ once we impose an additional block diagonal constraint $F'\in S$ to the nonnegative $F'$ ($S$ in Definition \ref{def:3}), in particular, $\Pi F=\left[{\rm Diag}(I_{k_1},\dots,I_{k_k});{\rm Diag}(F'_1,\dots,F'_k)\right]$, where $F'_i\in\mathbb R_{+}^{(n_i-k_i)\times k_i}$, $\sum_{i=1}^k n_i=n$, and $\Pi$ is a row-permutation matrix. So the $n_i$ points in cluster $i$ have coefficient matrix $F_i$ in the subspace spanned by $k_i$ ``anchors'' associated with $I_{k_i}$ in $\Pi F$, i.e., $\forall i\in[k], X_i=F_iX_{A_i}, \Pi_i F_i=\left[
                          \begin{array}{c}
                            I_{k_i} \\
                            F'_i \\
                          \end{array}
                        \right]$, where $X_i\in\mathbb R^{n_i\times p}$ are the points in cluster $i$, and $A=\biguplus_{i=1}^k A_i$, the disjoint union of all the anchors from the $k$ clusters. Hence, given clustering labels, our method is equivalent to applying a minimum conical hull problem model to points in each cluster. 
                        
                        Our goal is to find out the $k$ groups of anchors and simultaneously separate the points covered inside the associated $k$ cones $\{cone(X_{A_i})\}_{i\in[k]}$. In other words, we reduce SC to a ``cone clustering'' problem separate $k$ cones of data points. Each cone covers all the points in one cluster, and the faces spanned by its anchors separate these points from those covered by other cones. 
                        
                        Separable NMF is a special case of separable SC when $k=1$. When $k>1$, the cone clustering problem might be difficult to solve in high-dim space. But by applying the DCA scheme proposed in the next section, we will show that this problem can be reduced to several cone clustering problems on low-D hyperplanes, each can be solved by efficient existing clustering method.


} {See~\cite{nips2014supplemental} for an example of reducing the subspace clustering to general minimum conical hull problem.} 

\subsection{Latent Variable Model}\label{sec:32}

Different from deterministic MF, we build a system of equations from the moments of probabilistic latent variable models, and then formulate it as a general minimum conical hull problem, rather than directly solve it. Let the generalization model be $h\sim p(h;\alpha)$ and $x\sim p(x|h;\theta)$, where $h$ is a latent variable, $x$ stands for observation, and $\{\alpha, \theta\}$ are parameters. In a variety of graphical models such as GMMs and HMMs, we need to model conditional independence between groups of features. This is also known as the \textbf{multi-view assumption}. \emph{W.l.o.g.}, we assume that $x$ is composed of three groups(views) of features $\{x_i\}_{i\in[3]}$ such that $\forall i\neq j, x_i\independent x_j|h$. \textbf{We further assume the dimension $k$ of $h$ is smaller than $p_i$}, the dimension of $x_i$. Since the goal is learning $\{\alpha, \theta\}$, decomposing the moments of $x$ rather than the data matrix $X$ can help us get rid of the latent variable $h$ and thus avoid alternating minimization between $\{\alpha, \theta\}$ and $h$. When $\mathbb E(x_i|h) = h^TO_i^T$ (\textbf{linearity assumption}), the second and third order moments can be written in the form of matrix operator.
\begin{equation}\label{equ: moments23}
\left\{
  \begin{array}{ll}
    \mathbb E\left(x_i\otimes x_j\right)=\mathbb E[\mathbb E(x_i|h)\otimes \mathbb E(x_j|h)] = O_i\mathbb E(h\otimes h)O_j^T,  \\
    \mathbb E\left(x_i\otimes x_j \cdot\langle \eta,x_l\rangle\right)=O_i \left[\mathbb E(h\otimes h\otimes h)\times_3 (O_l\eta)\right]O_j^T, 
  \end{array}
\right.
\end{equation}
where $A \times_n U$ denotes the $n$-mode product of a tensor $A$ by a matrix $U$, $\otimes$ is the outer product, and the operator parameter $\eta$ can be any vector. We will mainly focus on the models in which $\{\alpha, \theta\}$ can be exactly recovered from conditional mean vectors $\{O_i\}_{i\in[3]}$ and $\mathbb E(h\otimes h)$\footnote{Note our method can also handle more complex models that violate the linearity assumption and need higher order moments for parameter estimation. By replacing $x_i$ in (\ref{equ: moments23}) with ${\rm vec}(x_i\otimes^n)$, the vectorization of the $n^{th}$ tensor power of $x_i$, $O_i$ can contain $n^{th}$ order moments for $p(x_i|h;\theta)$. However, since higher order moments are either not necessary or difficult to estimate due to high sample complexity, we will not study them in this paper.}, because they cover most popular models such as GMM and HMM in real applications.

The left hand sides (LHS) of both equations in (\ref{equ: moments23}) can be directly estimated from training data, while their right hand sides (RHS) can be written in a unified matrix form $O_iDO_j^T$ with $O_i\in\mathbb R^{p_i\times k}$ and $D\in\mathbb R^{k\times k}$. By using different $\eta$, we can obtain $2\leq q\leq p_l+1$ independent equations, which compose a system of equations for $O_i$ and $O_j$. Given the LHS, we can obtain the column spaces of $O_i$ and $O_j$, which respectively equal to the column and row space of $O_iDO_j^T$, a low-rank matrix when $p_i>k$. In order to further determine $O_i$ and $O_j$, our discussion falls into two types of $D$.

{\bf When $D$ is a diagonal matrix.} This happens when $\forall i\neq j, \mathbb E(h_ih_j)=0$. A common example is that $h$ is a label/state indicator such that $h=e_i$ for class/state $i$, e.g., $h$ in GMM and HMM. In this case, the two $D$ matrices in RHS of (\ref{equ: moments23}) are
\begin{equation}\label{equ: Dmatrix}
\left\{
  \begin{array}{ll}
    \mathbb E(h\otimes h)={\rm Diag}(\overrightarrow{\mathbb E(h^2_i)}),  \\
    \mathbb E(h\otimes h\otimes h)\times_3 (O_l\eta)={\rm Diag}(\overrightarrow{\mathbb E(h^3_i)}\cdot O_l\eta),
  \end{array}
\right.
\end{equation}
where $\overrightarrow{\mathbb E(h^t_i)}=\left[\mathbb E(h^t_1),\dots,\mathbb E(h^t_k)\right]$. So either matrix in LHS of (\ref{equ: moments23}) can be written as a sum of $k$ rank-one matrices, i.e.,  $\sum_{s=1}^k\sigma^{(s)}O_i^s\otimes O_j^s$, where $O_i^s$ is the $s^{th}$ column of $O_i$.

The general separability assumption posits that the set of $k$ rank-one basis matrices constructing the RHS of (\ref{equ: moments23}) is a unique subset $A\subseteq[n]$ of the $n$ samples of $x_i\otimes x_j$ constructing the left hand sides, i.e., $O_i^s\otimes O_j^s=[x_i\otimes x_j]_{A_s}=X_{A_s,i}\otimes X_{A_s,j}$, the outer product of $x_i$ and $x_j$ in $(A_s)^{th}$ data point. Therefore, by applying $q-1$ different $\eta$ to (\ref{equ: moments23}), we obtain the system of $q$ equations in the following form, where $Y^{t}$ is the estimate of the LHS of $t^{th}$ equation from training data.
\begin{equation}\label{equ:vec}
\forall t\in[q], Y^{(t)}=\sum_{s=1}^k\sigma_{t,s}[x_i\otimes x_j]_{A_s}\Leftrightarrow[{\rm vec}(Y^{(t)})]_{t\in[q]}=\sigma[{\rm vec}(X_{t,i}\otimes X_{t,j})]_{t\in A}.
\end{equation}
The right equation in (\ref{equ:vec}) is an equivalent matrix representation of the left one. Its LHS is a $q\times p_ip_j$ matrix, and its RHS is the product of a $q\times k$ matrix $\sigma$ and a $k\times p_ip_j$ matrix. By letting $X\leftarrow[{\rm vec}(Y^{(t)})]_{t\in[q]}$, $F\leftarrow\sigma$ and $Y\leftarrow[{\rm vec}(X_{t,i}\otimes X_{t,j})]_{t\in[n]}$, we can fit (\ref{equ:vec}) to $X=FY_A$ in Definition \ref{def:3}. Therefore, learning $\{O_i\}_{i\in[3]}$ is reduced to selecting $k$ rank-one matrices from $\{X_{t,i}\otimes X_{t,j}\}_{t\in[n]}$ indexed by $A$, and defining the extreme rays of a conical hull covering the $q$ matrices $\{Y^{(t)}\}_{t\in[q]}$. Given the anchor set $A$, we have $\hat O_i=X_{A,i}$ and $\hat O_j=X_{A,j}$ by assigning real data points indexed by $A$ to the columns of $O_i$ and $O_j$. Given $O_i$ and $O_j$, $\sigma$ can be estimated by solving (\ref{equ:vec}). In many models, a few rows of $\sigma$ are sufficient to recover $\alpha$.

\arxivalt{
In (\ref{equ:vec}), the dimension of each matrix after vectorization is $p_ip_j$. In practice, this could lead to computational burden. Moreover, real data could suffers from missing features, which make computing all entries of $X_{t,i}\otimes X_{t,j}$ impossible. However, thanks to the matrix completion research \cite{MatrixCompletion,CoherentMC}, when $k\ll\min\{p_i,p_j\}$, we can retain merely $m=\mathcal O(\max\{p_i, p_j\}k\log^2(p_i+p_j))\ll p_ip_j$
entries in the vectorization of each matrix from $\{Y^{(t)}\}_{t\in[q]}$ and $\{X_{t,i}\otimes X_{t,j}\}_{t\in[n]}$ in (\ref{equ:vec}). \emph{W.h.p.}, the true $A$ can still be successfully recovered from such partial information.}{See~\cite{nips2014supplemental} for a practical acceleration trick based on matrix completion.}

{\bf When $D$ is a symmetric matrix with nonzero off-diagonal entries.} This happens in ``admixture'' models, e.g., $h$ can be a general binary vector $h\in\{0,1\}^k$ or a vector on the probability simplex, and the conditional mean $\mathbb E(x_i|h)$ is a mixture of columns in $O_i$. The most well known example is LDA, in which each document is generated by multiple topics. 


We apply the general separability assumption by only using the first equation in (\ref{equ: moments23}), and treating the matrix in its LHS as $X$ in $X=FX_A$. When the data are extremely sparse, which is common in text data, selecting the rows of the denser second order moment as bases is a more reasonable and effective assumption compared to sparse data points. In this case, the $p$ rows of $F$ contain $k$ unit vectors $\{e_i\}_{i\in[k]}$. This leads to a natural assumption of ``anchor word'' for LDA \cite{AroraLDA}. 

When the data is not sure to be sparse, as we will show in the example of Kalman filter, we apply a ``bilateral separability assumption'' to both matrices on LHS of (\ref{equ: moments23}). It is weaker than the one used for diagonal $D$ case, but much stronger than the one used for above LDA case. In particular, we apply the general separability assumption $X=FY_A$ to both the matrix on LHS of each equation in (\ref{equ: moments23}) and its transpose, i.e., the $k$ column of $O_i$ are selected from $n$ instances of $x_i$, while the $k$ columns of $O_j$ are selected from $n$ instances of $x_j$. In theory, it is hard to analyze the uniqueness of $O_i$ and $O_j$ under mild condition in this case, but it works pretty well in practice, and usually does ensure unique solution.

\arxivalt{
\subsection{Example: Multi-view Mixture Model}\label{sec:MM}

Mixture model (MM) $p(x)=\sum_{j=1}^k \sigma_jp(x; \theta^j)$ is a latent variable model broadly used in unsupervised learning including clustering, where $k$ is the number of clusters and is normally much less than the number of features $p$ in $x$. \emph{w.l.o.g.}, we assume the number of views to be $3$. Under multi-view assumption $x=\{x_i\}_{i\in [3]}$ with $\forall i\neq j, x_i\independent x_j|h$, MM generates an observation $x$ by firstly drawing a label indicator $h\sim \sigma\in\Delta^{k-1}$ from $\{e_i\}_{i\in[k]}$ and then drawing features of different views $x_i\sim p(x_i|\sum_{j=1}^kh_j\theta_i^j)$ independently. When $h=e_j$, i.e., $x$ belongs to class/cluster $j$, $x_i$ is drawn from $p(x_i|\theta_i^j)$, where $\theta_i^j$ is the distribution parameter for view $i$ in cluster $j$. We mainly focus on recovering the mean of $p(x_i|\theta_i^j)$ for all $\{i,j\}$\footnote{However, as we mentioned below (\ref{equ: moments23}), it is possible to recover covariance or higher moments for each $p(x_i|\theta_i^j)$ by using higher order sample moments of $x_i$.}, which is the learning goal of a majority number of mixture models in practice.

One example of MM is Gaussian MM (GMM), in which $p(x_i|\theta_i^j)$ is Gaussian $\mathcal N((O_i^j)^T, \Sigma_i^j)$, where $O_i^j$ is the $j^{th}$ column of $O_i\in\mathbb R^{p_i\times k}$. Thus we have $\mathbb E(x_i|h)=h^TO_i^T$, which is consistent to the linearity assumption. This GMM exactly fits the latent variable model we presented in \S\cref{sec:32}. Since $h\in\{e_i\}_{i\in[k]}$, $\mathbb E(h_ih_j)=0$, which implies the two $D$ matrices are diagonal. So GMM falls into the first type of latent variable model in \S\cref{sec:32}. By (\ref{equ: moments23}) and (\ref{equ:vec}), given data matrix $X$ where $X_{t,i}$ is view $i$ of data point $t$ and is the $t^{th}$ row of $X_i$, learning GMM can be reduced to solving a general conical hull problem in the form of $X=FY_A$ by letting 
\begin{equation}\label{equ:XYinGMM}
X\leftarrow\frac{1}{n}\left[{\rm vec}[X_1^TX_2]; {\rm vec}[X_1^T{\rm Diag}(X_3\eta_t)X_2]_{t\in[q]}\right], Y\leftarrow[{\rm vec}(X_{t,1}\otimes X_{t,2})]_{t\in[n]}.
\end{equation}

In practical algorithm, we can further apply the ``matrix completion'' trick at the end of \S\cref{sec:32} to the columns of $X$ and $Y$ in (\ref{equ:XYinGMM}) by randomly sampling a subset $\Omega\subseteq [p_1]\times [p_2]$, i.e.,
\begin{equation}\label{equ:XYinGMM1}
X\leftarrow\frac{1}{n}\left[{\rm vec}[(X_1^TX_2)_\Omega]; {\rm vec}[(X_1^T{\rm Diag}(X_3\eta_t)X_2)_\Omega]_{t\in[q]}\right], Y\leftarrow[{\rm vec}((X_{t,1}\otimes X_{t,2}))_\Omega]_{t\in[n]}.
\end{equation}
The identifiability of $A$ in Lemma \ref{lemma:iden} still holds \emph{w.h.p.} in this case. However, the resulting model requires much less computations than the ``full matrix'' model.

Empirically (and shown in \S\cref{sec:5}), we find out that our method performs appealingly even when applying the $3$-view assumption to data that do not have multi-view features by randomly splitting all features into $3$ groups. Obviously, some feature correlation information is ignored in this case, but the rest correlations between views captured by our method are usually sufficient to produce a reliable estimation of the parameters. In addition, the Gaussian distribution assumption is only a special case satisfying the linearity assumption $\mathbb E(x_i|h)=h^TO_i^T$ made in our model. So DCA can be actually applied to more general mixture models. 

\subsection{Example: Hidden Markov Model}\label{sec:HMM}

Hidden Markov model (HMM) is a latent variable model broadly used to analyze sequential data and time series. HMM can be depicted as a Markov chain of hidden states $\{h_t\}_{t\in[T]}$ ($h_t=e_i$ if the state is $i$), each $h^t$ generates the observation $x_t$ at time $t$. Markov chain property implies $\forall j\neq t, x_t\independent x_j|h_t$, so multi-view assumption holds. The generalization process is $h_1\sim\pi\in\Delta^{k-1}$, $h_t\sim  T{h_{t-1}}\in\Delta^{k-1}$ and $x_t\sim p(x_t|h_t, O)$, where $T\in\mathbb R^{k\times k}$ is the transition matrix such that $T_{i,j}=p(h_t=e_i|h^{t-1}=e_j)$, and $O\in\mathbb R^{p\times k}$ is the emission matrix such that $\mathbb E(x_t|h_t)=h_t^TO^T$. This is in consistency with our linearity assumption. Normally, the number of states $k\ll p$, the dimension of observation $x$. Given one or several sequences of observations $\{x_t\}_{t\in[T]}$, the goal of learning HMM is to estimate parameters $\{O, T\}$.

HMM can be converted to a special case of MM by integrating out the two hidden states before and after the current one. In particular, \emph{w.l.o.g.}, for each triple of observations $\{x_1,x_2,x_3\}$ and the corresponding hidden states $\{h_1,h_2,h_3\}$, let $h_1\sim\pi$, by integrating out $h_1$ and $h_2$, we have \cite{SpectralMM}  
\begin{equation}\label{equ:hmm2gmm}
\left\{
  \begin{array}{ll}
   \mathbb E(x_2|h_2)=h_2^TO^T\rightarrow h_2^TO_2^T, h_2\sim T\pi,\\
   \mathbb E(x_1|h_2)=\sum_{h_1}\mathbb E(x_1|h_1)\cdot  \frac{p(h_2|h_1)p(h_1)}{p(h_2)}=h_2^T\left[O{\rm Diag}(\pi)T^T{\rm Diag}((T\pi)^{-1})\right]^T\rightarrow h_2^TO_1^T,\\
   \mathbb E(x_3|h_2)=\sum_{h_3}\mathbb E(x_3|h_3)p(h_3|h_2)=h_2^T[OT]^T\rightarrow h_2^TO_3^T.
  \end{array}
\right.
\end{equation}
Therefore, we can obtain a system of equations in the same form of (\ref{equ: moments23}) with diagonal $D$, i.e.,
\begin{equation}\label{equ:hmm}
\left\{
  \begin{array}{ll}
    \mathbb E\left(x_2\otimes x_3\right)=O{\rm Diag}(T\pi)[OT]^T,  \\
    \mathbb E\left(x_2\otimes x_3 \cdot\langle \eta,x_1\rangle\right)=O{\rm Diag}(T\pi\cdot O_1\eta)[OT]^T, 
  \end{array}
\right.
\end{equation}
A data matrix $X=[X_1,X_2,X_3]$ with $X_i\in\mathbb R^{n\times p}$ whose rows are all the triples $\{x_1,x_2,x_3\}$ can be built from available sequences of observations. General separability assumption posits that the conditional means $\mathbb E(x_2|h_2)$ and $\mathbb E(x_3|h_2)$, i.e., the columns in $O$ and $OT$, are selected from real instances of observations. So we can fit the problem of learning emission matrix $O$ in HMM to a conical hull problem $X=FY_A$ by using the same formulas (\ref{equ:XYinGMM}) or (\ref{equ:XYinGMM1}) for GMM.

We will show how the transition matrix $T$ can be immediately recovered given the anchor set $A$ in \S\cref{sec:4}.

\subsection{Example: Kalman Filter}

When extending the discrete latent state $h$ in HMM to more general continuous latent variable, we can obtain a linear dynamical system (LDS), which has been widely used in filtering and smoothing of time series data.
\begin{equation}\label{equ:LDS}
\begin{array}{ll}
&h_{t}=Th_{t-1}+w_t,\\
&x_t=h_t^TO^T+v_t,
\end{array}
\end{equation}
where $h_t$ and $x_t$ are hidden (continuous) state and the associated observation (output) respectively at time $t$, while $w_t$ and $v_t$ are noise terms independent to $Th_t$ and $Oh_t$ respectively. Hence, when distributions $p(w_t)$ and $p(v_t)$ are symmetric, we have linearity $\mathbb E(h_t|h_{t-1})=Th_{t-1}$ and $\mathbb E(x_t|h_t)=h_t^TO^T$, which are analogous to HMM. Similar to HMM, the multi-view or conditional independence assumption automatically holds for such Markov typed model, and it is also reasonable to assume that $k\ll p$. The goal of learning is to estimate $\{O, T\}$ and the distribution parameters $\Theta$ for $p(w_t)$ and $p(v_t)$.

Compared to HMM, the only difference here is that $h$ changes from disjoint discrete state to $k$-D continuous variable. In our approach, this will lead to different latent variable moments and thus discrepant matrix $D$ in \S\cref{sec:32}. Therefore, reducing (\ref{equ:LDS}) to the general minimum conical hull problem exactly follows the same procedures for HMM in \S\cref{sec:HMM} except replacing $h_2\sim T\pi$ to $h_2\sim p(Th_1 + w_2)$, i.e.,
\begin{equation}\label{equ:kf2hmm}
\left\{
  \begin{array}{ll}
   \mathbb E(x_2|h_2)=h_2^TO^T\rightarrow h_2^TO_2^T, h_2\sim p(Th_1 + w_2),\\
   \mathbb E(x_1|h_2)=\sum_{h_1}\mathbb E(x_1|h_1)\cdot  \frac{p(h_2|h_1)p(h_1)}{p(h_2)}=h_2^T\left[O{\rm Diag}(\pi)T^T{\rm Diag}((T\pi)^{-1})\right]^T\rightarrow h_2^TO_1^T,\\
   \mathbb E(x_3|h_2)=\sum_{h_3}\mathbb E(x_3|h_3)p(h_3|h_2)=h_2^T[OT]^T\rightarrow h_2^TO_3^T.
  \end{array}
\right.
\end{equation}
Therefore, we can obtain a system of equations in the same form of (\ref{equ: moments23}), i.e.,
\begin{equation}\label{equ:kf}
\left\{
  \begin{array}{ll}
    \mathbb E\left(x_2\otimes x_3\right)=O\mathbb E(h_2\otimes h_2)[OT]^T,  \\
    \mathbb E\left(x_2\otimes x_3 \cdot\langle \eta,x_1\rangle\right)=O\mathbb [E(h_2\otimes h_2\otimes h_2)\times_3 (O_1\eta)][OT]^T, 
  \end{array}
\right.
\end{equation}
In most cases, the square matrix $D$ in both equations given in (\ref{equ:kf}) is hardly to be diagonal, even when both $\mathbb E(h_1\otimes h_1)$ and $\mathbb E(w_2\otimes w_2)$ are diagonal. This falls into the second type of $D$ discussed in \S\cref{sec:HMM}. However, since the observations are usually not sparse, different from LDA in \S\cref{sec:LDA} where we directly apply original separability assumption to the moment matrices, we apply a ``bilateral separability assumption'' to both matrices on LHS of (\ref{equ:kf}). It is weaker than the one used for diagonal $D$ case, but much stronger than the one used for LDA case. In particular, we apply the general separability assumption to both the matrix on LHS of each equation in (\ref{equ:kf}) and its transpose, i.e., the $k$ column of $O$ are selected from $n$ instances of $x_2$, while the $k$ columns of $OT$ are selected from $n$ instances of $x_3$. To fit the notations in $X=FY_A$, let
\begin{equation}
\left\{
  \begin{array}{ll}
  X\leftarrow \frac{1}{n}\left[X_2^TX_3; [X_2^T{\rm Diag}(X_1\eta_t)X_3]_{t\in[q]}\right], Y\leftarrow X_2,\\
  X\leftarrow \frac{1}{n}\left[X_2^TX_3; [X_2^T{\rm Diag}(X_1\eta_t)X_3]_{t\in[q]}\right]^T, Y\leftarrow X_3.\\
  \end{array}
\right.
\end{equation}
After achieving $O$ and $OT$, square matrix $T$ can be immediately determined by solving a linear equation.

When the noise terms in this system are randomly drawn from Gaussians, (\ref{equ:LDS}) leads to the infamous Kalman filter model.

\subsection{Example: Latent Dirichlet Allocation}\label{sec:LDA}

Latent Dirichlet Allocation (LDA) is a latent variable model that is widely applied to bag-of-words features for text and vision data in order to extract semantic topics, whose effectiveness has been proved in clustering and classification tasks. It generates the $j^{th}$ word $x_j\in\{e_i\}_{i\in[p]}$ in a document by firstly drawing a topic proportion $h\sim {\rm Dir}(\alpha)$ (Dirichlet distribution with parameter $\alpha=\{\alpha_i\}_{i\in[k]}$) for the document, then drawing a topic $z_j\sim h$ from $\{e_i\}_{i\in[k]}$ with associated probability $\beta=z_j^TO^T$ over $p$ words in vocabulary, and drawing word $x_j\sim\beta$ at last. The topic probability matrix $O$ stores the conditional probabilities $O_{i,j}=p(x=e_i|z=e_j)$.

Given $h$, different words in a document are generated independently, so the multi-view assumption $\forall j\neq t, x_t\independent x_j|h$ holds. Since $\mathbb E(x_j|h)=\sum_{z}\mathbb E(x_j|z)p(z|h)=h^TO^T$, the linearity assumption is satisfied on LDA. Normally, the number of topics $k\ll p$, the number of words in vocabulary. Given the bag-of-words features of $n$ documents $X\in\mathbb Z_+^{n\times p}$, where $X_{i,j}$ denotes the number of times the $j^{th}$ word appears in the $i^{th}$ document, the goal of learning LDA is to estimate $O$ and $\alpha$.

We consider the two-gram statistics, \emph{w.l.o.g.}, the word-word co-occurrence matrix for the first two words $x_1$ and $x_2$ in a document, i.e., 
\begin{equation}\label{equ:lda2order}
\mathbb E(x_1\otimes x_2)=\mathbb E\left[\mathbb E(x_1|h)\otimes \mathbb E(x_2|h)\right]=O\mathbb E(h\otimes h)O^T,
\end{equation}
where $\mathbb E(h\otimes h)$ is the topic-topic covariance matrix, which is exactly the $D$ matrix for the first equation in (\ref{equ: moments23}). Since LDA is an ``admixture'' model that allows multiple topics in one document, this $D$ matrix has nonzero off-diagonal entries, and thus LDA falls into the second type of latent variable model studied in \S\cref{sec:32}. Therefore, under separability assumption, learning $O$ in LDA is reduced to conical hull problem $X=FX_A$, where $X$ here is the word-word co-occurrence matrix in (\ref{equ:lda2order}), and $O$ is $F$ after column normalization (to the probability simplex). In order to take advantage of all word pairs rather than just the first one, the word-word co-occurrence matrix $\mathbb E(x_1\otimes x_2)$ is estimated by averaging over all word pairs in all documents, i.e., the $X$ in $X=FX_A$ can be estimated from the bag-of-words feature matrix $X$ as
\begin{equation}\label{equ:H2Q}
X\leftarrow\bar X^T\bar X-{\rm Diag}\left(\mathbf{1}^T\hat X\right),\hat X_{i,j}\leftarrow\frac{X_{i,j}}{m_i(m_i-1)},\bar X_{i,j}\leftarrow\frac{X_{i,j}}{\sqrt{m_i(m_i-1)}}, m\leftarrow X\mathbf{1}.
\end{equation}

We will show how the parameter $O$ and $\alpha$ can be immediately recovered given the anchor set $A$ in \S\cref{sec:4}.

}{See~\cite{nips2014supplemental} for the example of reducing multi-view mixture model, HMM, and LDA to general minimum conical hull problem. It is also worthy noting that for the special case of LDA, we can prove the equivalence between our method and a Bayesian learning method~\cite{AroraLDA}, but our method results in a faster algorithm, see Theorem 4 in~\cite{nips2014supplemental}.} 

\vspace{-3mm}
\section{Algorithms for Minimum Conical Hull Problem}\label{sec:4}
\vspace{-2mm}
%

\subsection{Divide-and-Conquer Anchoring (DCA) for General Minimum Conical Hull Problem}

\arxiv{
According to the above section, parameter learning of MF and latent variable model can be reduced to finding the anchor set $A$ for a conical hull such that $X=FY_A$. 
In this section, we will focus on a novel distributed learning scheme that spans different dimensions in developing algorithms for $X=FY_A$. The proposed divide-and-conquer anchoring (DCA) in Algorithm \ref{A:DCA} decomposes the conical hull problem to multiple (much easier) sub-problems in extremely low dimensions, which can be solved in parallel. The first DCA algorithm was proposed in \cite{DCANMF} only for separable NMF, but this paper will largely extend the idea to much richer class of problems and algorithm designs.

} The key insights of DCA come from two observations on the geometry of the convex cone. First, projecting a conical hull to a lower-D hyperplane partially preserves its geometry. This enables us to distribute the original problem to a few much smaller sub-problems, each handled by a solver to minimum conical hull problem. Secondly, there exists an ultrafast anchoring algorithm for sub-problem on 2D plane, which only picks two anchor points based on their angles to an axis without iterative optimization or greedy pursuit. This results in a significantly efficient DCA algorithm that can be solely used, or embedded as a subroutine checking if a point is covered in a conical hull.
\vspace{-3mm}
\subsection{Distributing Conical Hull Problem to Sub-problems in Low Dimensions} 
\vspace{-1mm}
Due to the convexity of cone, a low-D projection of a conical hull is still a conical hull that covers the projections of the same points covered in the original conical hull, and generated by the projections of a subset of anchors on the extreme rays of the original conical hull. 

\begin{lemma}\label{lemma:proj}
For arbitrary point $x\in cone(Y_A)\subset R^p$, where $A$ is the index set of the $k$ anchors (generators) selected from $Y$, for any $\Phi\in\mathbb R^{p\times d}$ with $d\leq p$, we have
\begin{equation}
\exists \tilde A\subseteq A: x\Phi\in cone(Y_{\tilde A}\Phi),
\end{equation}
\end{lemma}

\vspace{-3mm}
Since merely a subset of $A$ remains as anchors after projection, solving a minimum conical hull problem on a single low-D hyperplane rarely returns all the anchors in $A$. However, the whole set $A$ can be recovered from the anchors detected on multiple low-D hyperplanes. 
By sampling the projection matrix $\Phi$ from a random ensemble $\mathbb M$, it can be proved that \emph{w.h.p.} solving only $s=\mathcal O(ck\log k)$ sub-problems are sufficient to find all anchors in $A$. \arxivalt {This is the bound for the worst case, i.e., the constant $c$ only changes with the probability that the most ``flat'' anchor on the conical hull surface is still an anchor on the low-D hyperplane after projection. However, it is interesting that when the anchor is too ``flat'', it is very close to the face spanned by its adjacent anchors, and thus the failure of detecting it still leads to a sufficiently similar conical hull. In other words, this property indicates robustness to losing unimportant anchors.} {See~\cite{nips2014supplemental} for the robustness of our method to the failure in identifying ``flat'' anchors.} 

\begin{algorithm}[htp]\small
\caption{DCA($X,Y,k,\mathbb M$)}
\begin{algorithmic}
\STATE \textbf{Input:}  Two sets of points (rows) $X\in\mathbb R^{n\times p}$ and $Y\in\mathbb R^{m\times p}$ in matrix forms (ref. Table \ref{table:models} to see $X$ and $Y$ for different models), number of latent factors/variables $k$, random matrix ensemble $\mathbb M$;
\STATE \textbf{Output:} Anchor set $A\subseteq [m]$ such that $\forall i\in[n],X_i\in cone(Y_A)$;
\STATE \emph{Divide Step (in parallel)}:
\FOR {$i=1\to s:=\mathcal O(k\log k)$}
\STATE Randomly draw a matrix $\Phi\in\mathbb R^{p\times d}$ from $\mathbb M$;
\STATE Solve sub-problem such as $\tilde A^t=\MCH(X\Phi, Y\Phi)$ by any solver, e.g., (\ref{equ:2Danchor});
\ENDFOR
\STATE \emph{Conquer Step}:
\STATE $\forall i\in [m]$, compute $\hat g(Y_i)=(1/s)\sum_{t=1}^s\mathds{1}_{\tilde A^t}(Y_i)$;
\STATE Return $A$ as index set of the $k$ points with the largest $\hat g(Y_i)$.
\end{algorithmic}\label{A:DCA}
\end{algorithm}\normalsize

For the special case of NMF when $X=FX_A$, the above result was proved in \cite{DCANMF}. However, the analysis cannot be trivially extended to the general conical hull problem when $X=FY_A$ (see \arxivalt{Left plot of Figure \ref{fig:mch}}{Figure \ref{fig:cone}}). A critical reason is that the converse of Lemma \ref{lemma:proj} does not hold: the uniqueness of the anchor set $\tilde A$ on low-D hyperplane could be violated, because non-anchors in $Y$ may have non-zero probability to be projected as low-D anchors. Fortunately, we can achieve a unique $\tilde A$ by defining a ``minimal conical hull'' on a low-D hyperplane. Then Proposition \ref{prop:mch} reveals when w.h.p such $\tilde A$ is a subset of $A$.

\notarxiv{
\begin{wrapfigure}{r}{0.3\linewidth}
\begin{center}\vspace{-12mm}
 \includegraphics[width=0.9\linewidth]{cproof.pdf}
\end{center}\vspace{-4mm}
   \caption{Proposition \ref{prop:mch}.}
\label{fig:cproof}
\vspace{-8mm}
\end{wrapfigure}
}

\arxiv{ 
\begin{figure}[htp]
\begin{center}
 \includegraphics[width=1\linewidth]{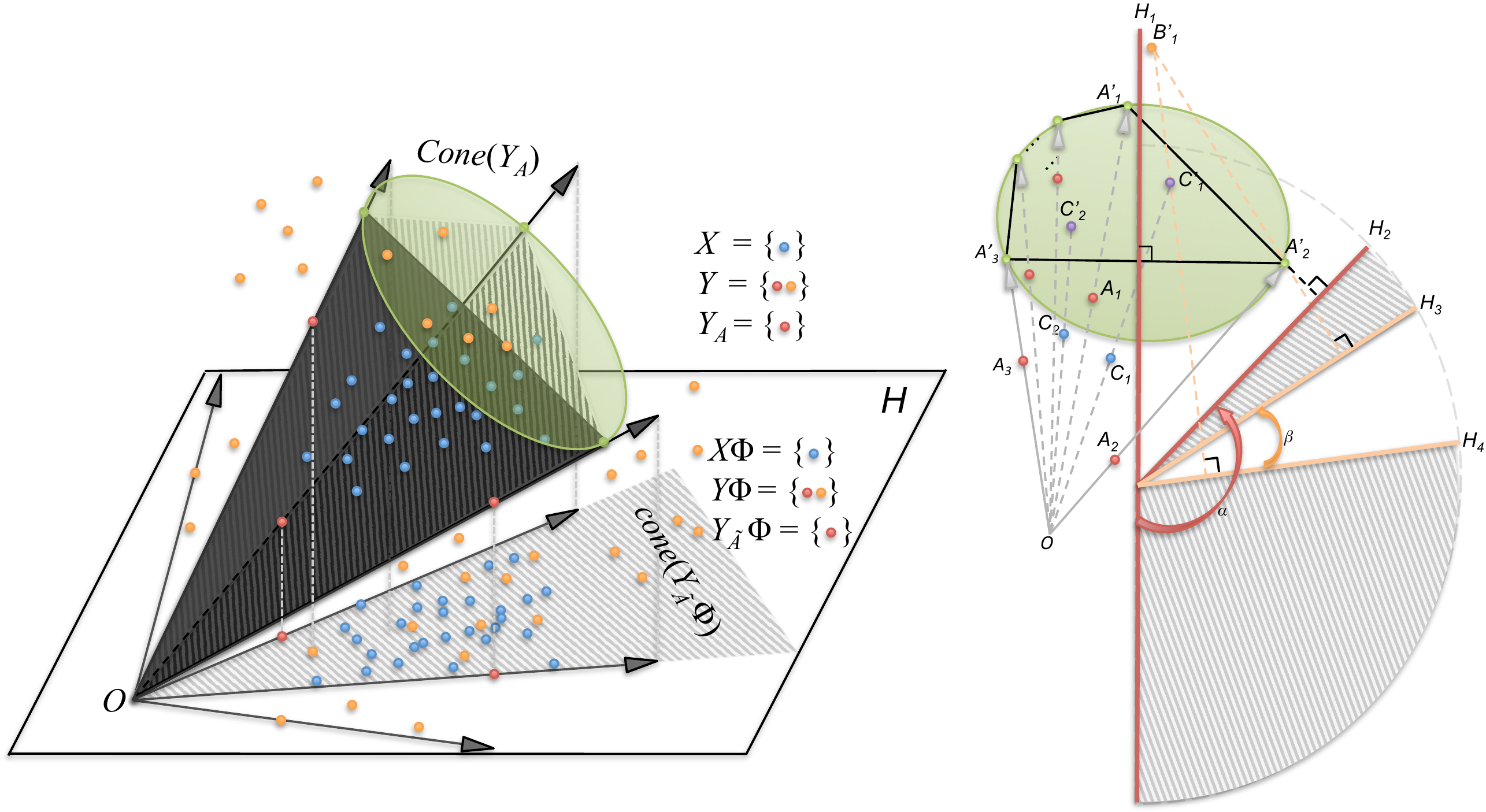}
\end{center}
   \caption{LEFT: Geometry of conical hull problem under generalized separability assumption. RIGHT: Illustration of proof to Proposition \ref{prop:mch}.}
\label{fig:mch}
\end{figure}
}

\begin{definition}[\bf{Minimal conical hull}]
\label{def:mch}
Given two sets of points(rows) $X$ and $Y$, the conical hull spanned by anchors (generators) $Y_A$ is the minimal conical hull covering all points in $X$ iff
\begin{equation}
\forall \{i,j,s\}\in\left\{i,j,s\mid i\in A^C=[m]\setminus A, j\in A, s\in [n], X_s\in cone(Y_A)\cap cone(Y_{i\cup (A\setminus j)})\right\}
\end{equation}
we have $\widehat{X_sY_i}>\widehat{X_sY_j}$, where $\widehat{xy}$ denotes the angle between two vectors $x$ and $y$. The solution of minimal conical hull is denoted by $A=\MCH(X,Y)$.
\end{definition}

It is easy to verify that the minimal conical hull is unique, and the general minimum conical hull problem $X=FY_A$ under general separability assumption (which leads to the identifiability of $A$) is a special case of $A=\MCH(X,Y)$. In DCA, on each low-D hyperplane $H_i$, the associated sub-problem aims to find the anchor set $\tilde A^i=\MCH(X\Phi^i, Y\Phi^i)$. The following proposition gives the probability of $\tilde A^i\subseteq A$ in a sub-problem solution.

\begin{proposition}[\bf{Probability of success in sub-problem}]
\label{prop:mch}
As defined in \arxivalt{the right plot of Figure \ref{fig:mch}}{Figure \ref{fig:cproof}}, $A_i\in A$ signifies an anchor point in $Y_A$, $C_i\in X$ signifies a point in $X\in\mathbb R^{n\times p}$, $B_i\in A^C$ signifies a non-anchor point in $Y\in\mathbb R^{m\times p}$, the green ellipse marks the intersection hyperplane between $cone(Y_A)$ and the unit sphere $\mathbb S^{p-1}$, the superscript $\cdot'$ denotes the projection of a point on the intersection hyperplane. Define $d$-dim ($d\leq p$) hyperplanes $\{H_i\}_{i\in[4]}$ such that $A'_3A'_2\perp H_1, A'_1A'_2\perp H_2, B'_1A'_2\perp H_3, B'_1C'_1\perp H_4$, let $\alpha=\widehat{H_1H_2}$ be the angle between hyperplanes $H_1$ and $H_2$, $\beta=\widehat{H_3H_4}$ be the angle between $H_3$ and $H_4$. If $H$ with associated projection matrix $\Phi\in\mathbb R^{p\times d}$ is a $d$-dim hyperplane uniformly drawn from the Grassmannian manifold ${\rm Gr}(d,p)$, and $\tilde A=MCH(X\Phi, Y\Phi)$ is the solution of minimal conical hull problem $MCH(X\Phi, Y\Phi)$, we have
\begin{equation}
\Pr(B_1\in\tilde A)=\frac{\beta}{2\pi}, \Pr(A_2\in\tilde A)=\frac{\alpha-\beta}{2\pi}.
\end{equation}
\end{proposition}

\arxivalt {
\begin{proof}
This proposition can be immediately proved by using the right plot of Figure \ref{fig:mch}. When rotating $H$ from $H_1$ to $H_2$ by angle $\alpha$, $A_2$ will be identified as an anchor point of the minimal conical hull, except when $H$ is between $H_4$ and $H_3$, in which region the non-anchor point $B_1$ will be identified as an anchor point. Since the probability of being anchor point is proportional to the corresponding angle, we have $\Pr(B_1\in\tilde A)=\beta/(2\pi)$ and $\Pr(A_2\in\tilde A)=(\alpha-\beta)/(2\pi)$.

\end{proof}

It can be further verified that the angles $\alpha$ and $\beta$ can be computed from the data points shown in the plot such that
\begin{equation}
\alpha= \arccos\left(\frac{(A'_2-A'_1)^T(A'_2-A'_3)}{\|A'_2-A'_1\|\|A'_2-A'_3\|}\right),\beta = \arccos\left(\frac{(B'_1-A'_2)^T(B'_1-C'_1)}{\|B'_1-A'_2\|\|B'_1-C'_1\|}\right).
\end{equation}
Thus $\alpha$ and $\beta$ can be computed as constants for specific $\{A_1,A_2,A_3,B_1,C_1\}$. 

\textbf{Remarks:} 

It is obvious that a large $\Pr(A_2\in\tilde A)-\Pr(B_1\in\tilde A)=(\alpha-2\beta)/(2\pi)$ leads to a large probability for the success of a sub-problem in recovering a subset of $A$, i.e., $\tilde A\subseteq A$. 

The robustness to unimportant ``flat'' anchors still holds for this general model. When increasing an interior angle associated to vertex $A'_2$ of the polygon on the green intersection hyperplane, $A_2$ turns to be a ``flat'' anchor, and angle $\alpha$ in the right plot of Figure \ref{fig:mch} will become small, so does the probability of detecting $A_2$ as anchor on a low-D hyperplane. The above proof also indicates robustness to data noise. When a non-anchor point's projection on the intersection hyperplane $B'_i$ is close to the polygon defined by all vertexes $A'_i$, the probability of $B_i$ to be falsely identified as a anchor in a sub-problem will increase. But since $B_i$ is close to the true conical hull, such false non-anchor point still leads to a good approximation of the true conical hull. Therefore, even our analysis aims at precisely recovery of $A$ under noiseless assumption of data points, failure in recovering $A$ can still provide a good approximate of $cone(Y_A)$.

Although Proposition \ref{prop:mch} generally assumes that $\Phi$ is uniformly drawn from a Grassmannian manifold, empirically we can sample $\Phi$ from a rich class of random matrix ensembles $\mathbb M$, e.g., Gaussian random matrix ensemble, the ensemble composed of standard unit vectors $e_i$ (i.e., random feature selection) or real data vectors, and various sparse random matrix ensemble, which can bring evident acceleration to projection $X\Phi$ and $Y\Phi$.


According to Proposition \ref{prop:mch}, when $\alpha>2\beta$ for all tuples $\{A_1,A_2,A_3,B_1,C_1\}$, the $k$ points in $Y$ with the largest $\Pr(i\in\tilde A)$ compose the unique true anchor set $A$. Since $\Pr(i\in\tilde A)$ cannot be exactly known, DCA compares its unbiased estimator $\hat g(Y_i)=(1/s)\sum_{t=1}^s\mathds{1}_{\tilde A^t}(Y_i)$ of all points $Y_i$ in $Y$, where the number of sub-problems $s$ is the sample size of the estimator. Therefore, by using Chernoff bound, we can obtain the probability bound for the success of DCA in finding the true $A$.}{See~\cite{nips2014supplemental} for proof, discussion and analysis of robustness to unimportant ``flat'' anchors and data noise.}

\begin{theorem}[\bf{Probability bound}]
\label{th:prob}
Following the same notations in Proposition \ref{prop:mch}, suppose $p^{**}=\min_{\{A_1,A_2,A_3,B_1,C_1\}}(\alpha-2\beta)\ge c/k > 0$. It holds with probability at least $1- k \exp\left(-\frac{cs}{3k}\right)$ that DCA successfully identifies all the $k$ anchor points in $A$, where $s$ is the number of sub-problems solved in DCA.
\end{theorem}

\arxivalt {
\begin{proof}
We introduce two binary random variables $\xi_i^t=\mathds{1}_{\tilde A^t}(A_i)$ and $\kappa_j^t=\mathds{1}_{\tilde A^t}(B_j)$ indicating whether $A_i\in\tilde A^t$ and $B_j\in\tilde A^t$, respectively. According to Proposition \ref{prop:mch}, we have
\begin{equation}
\mathbb E(\xi^i_t)=\Pr(A_i\in\tilde A^t)=\alpha-\beta, \mathbb E(\kappa^j_t)=\Pr(B_j\in\tilde A^t)=\beta.
\end{equation}
DCA compares $\hat g(A_i)$ and $\hat g(B_j)$, i.e., 
\begin{equation}\label{equ:gAgB}
\hat g(A_i) = \frac{1}{s}\sum_{t=1}^s\xi_i^t, \hat g(B_j) = \frac{1}{s}\sum_{t=1}^s\kappa_j^t.
\end{equation}
The true anchor $A_i$ is identified as an anchor by DCA in the conquer step iff $\hat g(A_i)>\max\limits_{B_j}\hat g(B_j)$. By applying Chernoff bound to random variable $\hat g(A_i)-\hat g(B_j)$ (randomness is due to random hyperplane $H$), for any $\delta\in[0,1]$, $A_i$ and $B_j$, we have
\begin{equation}
\Pr \left(\sum_{t=1}^s(\xi_i^t-\kappa_j^t) < (1-\delta)sp^*_i\right)\le \exp\left(-\frac{\delta^2s p^*_i}{2+\delta}\right),
\end{equation}
where $p^*_i=\min\limits_{B_j}(\alpha-2\beta)$.

Let $\delta =1$ and $f(A_i)=\hat g(A_i)-\max\limits_{B_j}\hat g(B_j)$, by (\ref{equ:gAgB}), we have
\begin{align}
\Pr(f(A_i)=0)\le \exp\left(-\frac{s p^*_i}{3}\right).
\end{align}
This yields
\begin{equation}
\begin{array}{ll}
\Pr (\min\limits_{A_i\in A}f(A_i)>0) &= 1-\Pr(\cup_{A_i\in A} f(A_i) = 0)
\ge 1- \sum_{A_i\in A} \Pr(f(A_i)=0)\\
&\ge 1- \sum_{A_i\in A}  \exp\left(-\frac{s p^*_i}{3}\right)\ge 1- k \exp\left(-\frac{sp^{**}}{3}\right).
\end{array}
\end{equation}
Since $p^{**}=\min_{A_i}p^*_i\ge c/k$, this completes the proof.
\end{proof}}{See~\cite{nips2014supplemental} for proof.} Given Theorem \ref{th:prob}, we can immediately achieve the following corollary about the number of sub-problems that guarantee success of DCA in finding $A$.

\begin{corollary}[\bf{Number of sub-problems}]
With probability $1-\delta$, DCA can correctly recover the anchor set $A$ by solving $\Omega(\frac{3k}{c}\log\frac{k}{\delta})$ sub-problems.
\end{corollary}

\arxivalt{
When $k\ll p$, i.e., the number of latent variables/factors are much less than the features or dimension of data, which is common in many learning models, DCA can learn the model parameters by solving an extremely small number of sub-problems in parallel, no matter what solver chosen for sub-problem. Thus it provides a significantly efficient learning scheme.

\textbf{Remarks:} 

It is worth noting that although DCA uses random projection to reduce the problem size, it is different from the random projection methods based on \textit{Johnson-Lindenstrauss (JL) Lemma} \cite{JLlemma} or its variants. Because DCA allows to project the data into extremely low-D subspace in which JL lemma does not hold, but solving sub-problems on multiple times of such random projections can still recover the true solution \emph{w.h.p.} In contrast, the JL Lemma based methods have to project the data into a single yet much higher dimensional subspace to gain high probability in recovering the original solution. \textbf{The idea of divide-and-conquer randomization in DCA is more preferred in developing randomized algorithm}, because 1) the complexity of solving a sub-problem is usually super-linear in data dimension; and 2) parallelizing the sub-problems in DCA gives further speedup.}{See~\cite{nips2014supplemental} for the idea of divide-and-conquer randomization in DCA, and its advantage over \textit{Johnson-Lindenstrauss (JL) Lemma} based methods.}

\vspace{-2mm}
\subsection{Anchoring on 2D Plane}

\arxiv{DCA provides a fast unified distributed learning scheme that can invoke any minimal conical hull problem solver as subroutine to solve the sub-problems. Although there exists several solvers for $X=FX_A$ for NMF, most of them depend on expensive iterative algorithms derived from optimization or greedy pursuit. Moreover, there is rarely known algorithm addressing the general model $X=FY_A$.} Although DCA can invoke any solver for the sub-problem on any low-D hyperplane, an ultrafast solver for the 2D sub-problem always shows high accuracy in locating anchors when embedded into DCA. Its motivation comes from the geometry of conical hull on a 2D plane, which is a special case of a $d$-dim hyperplane $H$ in the sub-problem of DCA. It leads to a non-iterative algorithm for $A=\MCH(X,Y)$ on the 2D plane. It only requires computing $n+m$ cosine values, finding the min/max of the $n$ values, and comparing the remaining $m$ ones with the min/max value.

According to \arxivalt{the left plot of Figure \ref{fig:mch}}{Figure \ref{fig:cproof}}, the two anchors $Y_{\tilde A}\Phi$ on a 2D plane have the min/max (among points in $Y\Phi$ ) angle (to either axis) that is larger/smaller than all angles of points in $X\Phi$, respectively. This leads to the following closed form of $\tilde A$.
\begin{equation}\label{equ:2Danchor}
\tilde A = \{\arg\min_{i\in[m]}(\widehat{(Y_i\Phi)\varphi}-\max_{j\in[n]}\widehat{(X_j\Phi)\varphi})_+, \arg\min_{i\in[m]}(\min_{j\in[n]}\widehat{(X_j\Phi)\varphi}-\widehat{(Y_i\Phi)\varphi})_+\},
\end{equation}
where $(x)_+=x$ if $x\geq 0$ and $\infty$ otherwise, and $\varphi$ can be either the vertical or horizontal axis on a 2D plane. By plugging (\ref{equ:2Danchor}) in DCA as the solver for $s$ sub-problems on random 2D planes, we can obtain an extremely fast learning algorithm.

Note for the special case when $X=Y$, (\ref{equ:2Danchor}) degenerates to finding the two points in $X\Phi$ with the smallest and largest angles to an axis $\varphi$, i.e., $\tilde A=\{\arg\min_{i\in[n]}\widehat{(X_i\Phi)\varphi}, \arg\max_{i\in[n]}\widehat{(X_i\Phi)\varphi}\}$. This is used in matrix factorization and the latent variable model with nonzero off-diagonal $D$.

\arxivalt{

\subsection{DCA as Subroutine of Other Methods}

Within lots of algorithms finding conical hull or other problems, testing whether a point $X_i$ from $X$ is covered in the (minimal) conical hull of $Y$, or equivalently, if $X_i\in cone(Y)$, dominates the computation per step. In previous works, the testing needs to compute the conical combination coefficients by solving a linear programming. DCA provides a much faster off-the-shelf subroutine which can be easily invoked by other methods \cite{AGKM} to gain a significant acceleration.

In particular, the $t^{th}$ sub-problem in DCA turns to test if the 2D projection $X_i\Phi$ is covered in $cone(Y\Phi)$, this requires to compute
\begin{equation}
\epsilon_i^t = \mathds{1}(X_i\Phi<\min Y\Phi)+\mathds{1}(X_i\Phi>\max Y\Phi).
\end{equation}
The conquer step in DCA becomes
\begin{equation}
\left\{
  \begin{array}{ll}
    X_i\in cone(Y), &{\rm if}~~\sum_{t=1}^s\epsilon_i^t=0 \\
    X_i\not\in cone(Y), &{\rm if}~~\sum_{t=1}^s\epsilon_i^t>0, 
  \end{array}
\right.
\end{equation}

\subsection{Examples}

In this section, we will present five examples of using general conical hull problem in \S\cref{sec:3} and DCA in Algorithm \ref{A:DCA} to develop scalable novel learning algorithms for five popular latent variable and matrix factorization models. Although finding the anchor set $A$ plays a major role in all algorithms, in practice each one also needs extra  preprocessing/post-processing steps, which will be highlighted in the following.

\subsubsection{DCA for Multi-view Mixture Model}

Given $X$ and $Y$ in (\ref{equ:XYinGMM}) or (\ref{equ:XYinGMM1}) from \S\cref{sec:MM}, \textbf{applying $DCA(X,Y,k,\mathbb M)$ in Algorithm \ref{A:DCA} to the $X$ and $Y$ in GMM is able to find out the anchor set $A$ \emph{w.h.p.}, and thus $\hat O_1=X_{A,1}$, $\hat O_2=X_{A,2}$, and $\hat O_3=X_{A,3}$.} Therefore, DCA learns GMM by assigning $k$ real data instances to the mean vectors of the $k$ components. This is reasonable because we can usually find a real data instance sufficiently close to the true mean in each cluster when $n$ is large enough. This results in a more interpretable GMM because the centroid of each cluster is no longer an artificial averaging, but a representative real data instance. 

\subsubsection{DCA for Hidden Markov Model}

Given $X$ and $Y$ in (\ref{equ:XYinGMM}) or (\ref{equ:XYinGMM1}) from \S\cref{sec:MM}, and following notations in \S\cref{sec:HMM}, \textbf{after obtaining $\hat O=X_{A,2}$, $\hat {OT}=X_{A,3}$ by running $A=$DCA($X,Y,k,\mathbb M$), we can immediately recover transition matrix $T$ by solving linear equation $OT=X_{A,3}$ with simplex constraints to the columns of $T$}. Since $T$ is small $k\times k$ matrix, there are lots of standard solvers that can quickly attain $T$.

\subsubsection{DCA for Latent Dirichlet Allocation}

Given $X$ in (\ref{equ:H2Q}) from \S\cref{sec:LDA}, \textbf{after running $A=\DCA(X,X,k,\mathbb M)$, solving the system of linear equations $X=FX_A$ under constraints $\forall i\in[k], F_{A_i,A_i}=1, F_{A_i,t\in[k]\setminus A_i}=0$ and non-negativity $F_{i,j}\geq 0$ gives $F$. Since each column of $O$ is on a (topic) probability simplex $\Delta^{p-1}$, column-wise normalization of $F$ gives us the estimate to $O$, i.e., $\forall i\in[k], \hat O_i=F_i/\|F_i\|_1$. Then $\alpha$ can be recovered from $\mathbb E(h\otimes h)$, i.e., the covariance of ${\rm Dir}(\alpha)$ that is achieved by solving (\ref{equ:lda2order}) given $\hat O$.}

Recently, we surprisingly discover that the above procedure equals to a Bayes learning algorithm proposed in \cite{AroraLDA}, whose major idea is to solve an NMF under simplex constraint by a greedy pursuit typed algorithm. Thus it can be seen as a special case of our conical hull model.

Comparing to the specific greedy algorithm developed for LDA in \cite{AroraLDA}, our method provides a unified scheme that can reduce more general models besides LDA to a conical hull problem, and the proposed DCA leads to a significantly efficient algorithm. In addition, DCA has much faster speed and easier implementation. This is because 1) Limited by the simplex constraint in NMF, the Bayes learning method decomposes the row-normalized $X$ in (\ref{equ:H2Q}), thus it needs to additionally compute $p(w_2=j|w_1=i)$ and $p(w_1=i)$ for normalization, and $\sum_{i} p(z=e_t|x=e_i)p(x=e_i)$ after NMF; and 2)The greedy algorithm in \cite{AroraLDA} finds the anchors of convex hull is slower than DCA using parallel and randomized strategy.

The equivalence between \cite{AroraLDA} and our method can be established by the following theorem. In order to make the comparison clear, we map all the notations in our method to those used in \cite{AroraLDA} such that $Q\leftarrow\mathbb E(x_1\otimes x_2)$, $V\leftarrow A$, $A\leftarrow O^T$, $v_t$ denotes the index of the anchor word for topic $t$, $\bar A_{t,i}\leftarrow A_{t,i}/A_{t,v_t}$, $w_i=j\Leftrightarrow x_i=e_j$, $z_i=j\Leftrightarrow z_i=e_j$. Then we use the notations in \cite{AroraLDA} throughout the theorem and its proof below.

\begin{theorem}\label{t:equallda}
Solving the conical hull problem $Q=\bar A^TQ_V$ proposed in this paper for LDA equals to Bayes learning of $p(z_1=t|w_1=i)$ by NMF proposed in \cite{AroraLDA}, i.e.,
\begin{equation}\label{equ:equallda}
Q_{i,j} = \sum\limits_{t=1}^k \bar A_{t,i}Q_{v_t,j}\Leftrightarrow
\left\{
  \begin{array}{ll}
  p(w_2=j|w_1=i)=\sum\limits_{t=1}^k p(z_1=t|w_1=i)\cdot p(w_2=j|z_1=t),\\
  p(w_1=i|z_1=t)=\frac{p(z_1=t|w_1=i)p(w_1=i)}{\sum_{i^{'}} p(z_1=t|w_1=i^{'})p(w_1=i^{'})}.
  \end{array}
\right.
\end{equation}
\end{theorem}

\begin{proof}
From the perspective of Bayes learning, the basic law of conditional probability gives us
\begin{equation}\label{equ:Bayes}
p(w_2=j|w_1=i)=\sum\limits_{t=1}^k p(z_1=t|w_1=i)p(w_2=j|z_1=t).
\end{equation}
Since $A_{t,i}=p(z_1=t|w_1=i)$, given $p(w_2=j|w_1=i)$ and $p(w_2=j|z_1=t)$, solving the linear equations (\ref{equ:Bayes}) over all $\{i,j\}$ pairs gives the unique solution of $A$ when $k\leq p$. To relate this Bayes learning to our conical hull problem with form $X=FX_A$, multiplying both sides of (\ref{equ:Bayes}) by $p(w_1=i)$ yields
\begin{equation}\label{equ:BayesD1}
p(w_2=j|w_1=i)p(w_1=i)=\sum\limits_{t=1}^k \frac{p(z_1=t|w_1=i)p(w_1=i)}{p(w_1=v_t)}\cdot p(w_2=j|z_1=t)p(w_1=v_t).
\end{equation}
The definition of anchor word $p(z_1=t|w_1=v_t)=1$ leads to
\begin{equation}
p(w_2=j|w_1=v_t)=p(w_2=j|z_1=t)p(z_1=t|w_1=v_t)=p(w_2=j|z_1=t),
\end{equation}
so the equation in (\ref{equ:BayesD1}) equals to
\begin{equation}\label{equ:BayesD}
\begin{array}{ll}
& p(w_1=i,w_2=j)\\
&=\sum\limits_{t=1}^k \frac{p(z_1=t|w_1=i)p(w_1=i)}{p(w_1=v_t)}\cdot p(w_2=j|w_1=v_t)p(w_1=v_t)\\
&=\sum\limits_{t=1}^k \frac{p(z_1=t|w_1=i)p(w_1=i)}{p(z_1=t)}\cdot\frac{p(z_1=t)}{p(w_1=v_t)}\cdot p(w_1=v_t, w_2=j)\\
&=\sum\limits_{t=1}^k \frac{p(w_1=i|z_1=t)}{p(w_1=v_t|z_1=t)}\cdot p(w_1=v_t, w_2=j).
\end{array}
\end{equation}
The last equality is due to Bayes' rule and the definition of anchor word $p(z_1=t|w_1=v_t)=1$, i.e., 
\begin{equation}
\begin{array}{ll}
& \frac{p(z_1=t|w_1=i)p(w_1=i)}{p(z_1=t)}=p(w_1=i|z_1=t),\\
&\frac{p(z_1=t)}{p(w_1=v_t)}=\frac{p(z_1=t)}{p(z_1=t|w_1=v_t)p(w_1=v_t)}=\frac{1}{p(w_1=v_t|z_1=t)}.
\end{array}
\end{equation}
Substitute $Q_{i,j}=p(w_1=i,w_2=j)$ and $\bar A_{t,i}=A_{t,i}/A_{t,v_t}=p(w_1=i|z_1=t)/p(w_1=v_t|z_1=t)$ into the above equations (\ref{equ:BayesD}), we achieve the model in the same form as conical hull problem
\begin{equation}
Q_{i,j}=\sum_{t=1}^k \frac{A_{t,i}}{A_{t,v_t}}\cdot Q_{v_t,j}.
\end{equation}
Since $A$ can be uniquely recovered as row-normalized $\bar A$, and the above reasoning is reversible (due to all the equalities), the equivalence between conical hull problem and the Bayes learning \cite{AroraLDA} given in (\ref{equ:equallda}) holds.
\end{proof}

\subsubsection{DCA for Non-negative Matrix Factorization}

The conical hull problem for NMF uses model $X=FX_A$. DCA for NMF has been proposed  in \cite{DCANMF}.

\subsubsection{DCA for Subspace Clustering}

Comparing to existing SC algorithms relying on expensive lasso-type optimizations and spectral clustering requiring SVD, a slightly modified DCA is able to provide a significantly more efficient algorithm. \textbf{In particular, we change each sub-problem of $A=\DCA(X,X,K=\sum_{i=1}^kk_i,\mathbb M)$ to a separable SC of $X\Phi$ on a low-D hyperplane, which can be solved by any available separable SC solver. A simple but effective one is to sample $\Phi\in\mathbb R^{p\times 2}$, project $X$ to an 2D plane, apply mean shift clustering algorithm \cite{meanshift} to the $n$-array of angles $\widehat{(X_i\Phi)\varphi}$, and add to $\tilde A$ the two points with the maximal and minimal angle in each cluster.} The reason for using mean shift is 1) it is fast and provides highly reliable clustering result on 1D values; and 2) the number of clusters can be automatically determined in it. 

Following all notations in \S\cref{sec:SC}, after obtaining $A$, we have to extract the anchors $A_i$ in $A$ for different cluster $i$. We use the fact that two anchors in the same cluster must keep lying in the same cluster on any low-D hyperplane (but the converse does not hold). \textbf{Thus we can build a graph Laplacian from similarity matrix $G\in\mathbb R^{K\times K}$ such that $G_{i,j}=\#( X_{A(i)}\Phi$ and $X_{A(j)}\Phi$ in the same cluster$)$, and spectral clustering \cite{Ng01onspectral} is able to give us the $k$ clusters of anchors $\{A_i\}_{i\in[k]}$.}}{See~\cite{nips2014supplemental} for embedding DCA as a fast subroutine into other methods, and detailed off-the-shelf DCA algorithms of NMF, SC, GMM, HMM and LDA. A brief summary is in Table \ref{table:models}.}
\vspace{-3mm}
\section{Experiments}\label{sec:5}
\vspace{-1mm}
\arxivalt{
\subsection{DCA for Non-negative Matrix Factorization on Synthetic Data} 

The experimental comparison results are shown in Figure \ref{fig:NMF}. Greedy algorithms SPA, XRAY and SFO achieves the best accuracy and smallest recovery error when the noise level is above $0.2$, but XRAY and SFO are the slowest two. SPA is slightly faster but still much slower than DCA. DCA with different number of sub-problems shows slightly less accuracy and larger error than greedy algorithms, but the difference is acceptable. Considering its significant acceleration, DCA offers an advantageous trade-off. LP-test~\cite{AGKM} has the exact solution guarantee, but it is not robust to noise, and too slow in speed. Therefore, DCA provides a much faster and more practical NMF algorithm with comparable performance to the best ones.

\begin{figure}[htp]
\begin{center}
 \includegraphics[width=1\linewidth]{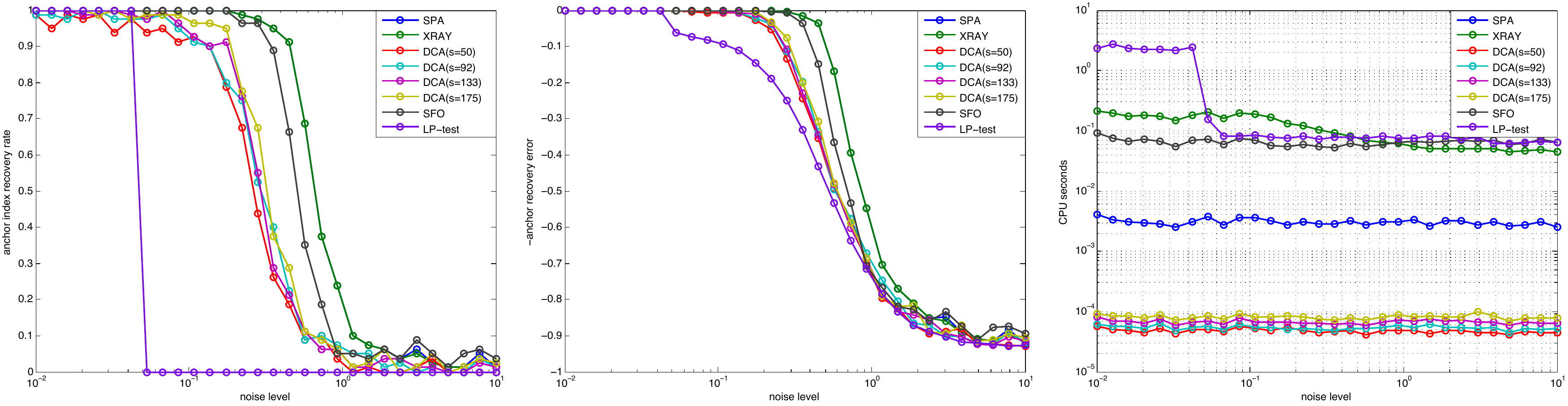}
\end{center}
   \caption{Separable NMF on randomly generated $300\times 500$ matrix, each point on each curve is the result by averaging $10$ independent random trials. SFO-greedy algorithm for submodular set cover problem. LP-test is the backward removal algorithm from~\cite{AGKM}. LEFT: Accuracy of anchor detection (higher is better). Middle: Negative relative $\ell_2$ recovery error of anchors (higher is better). Right: CPU seconds.}
\label{fig:NMF}
\end{figure}

\subsection{DCA for Gaussian Mixture Models on Synthetic Dataset}

\begin{figure}[htb]
\begin{center}
 \includegraphics[width=1\linewidth]{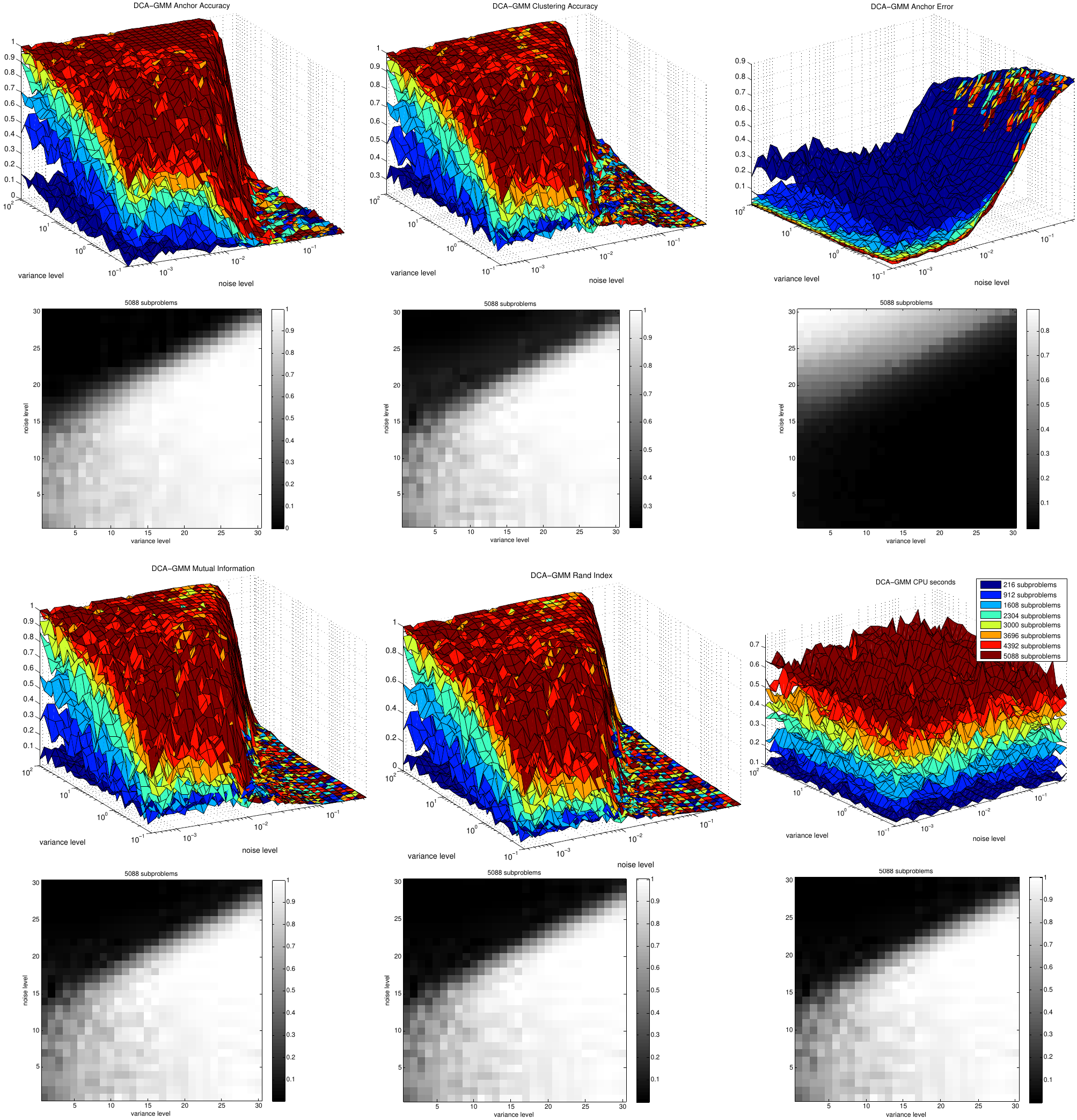}
\end{center}
   \caption{DCA-GMM on synthetic data. On a $30\times 30$ grid of different noise level and variance level, for each noise and variance pair, we randomly generate data of $k=5$ clusters with $300,500,400,300,500$ samples respectively, and $3$-view features of $200,120,160$ dimensions respectively. Mean vector of each cluster is added into the data as ``anchor'' we expect DCA to find out. In particular, each view of points in each cluster are drawn from a multivariate Gaussian distribution with the variance level, and then Gaussian noises of magnitude equal to the noise level are added to the points. We run DCA-GMM using different number of sub-problems, and report their performance by anchor accuracy, clustering accuracy, anchor error ($\ell_2$ relative recovery error), mutual information and rand index. The CPU seconds are reported too. Each point on each 3D plot is a result of averaging $10$ random trials in the same setting, and its height is the value of metric. Each 3D plot includes layers of surfaces associated with different number of sub-problems, we also report the top layer as a 2D plot below each 3D plot.}
\label{fig:GMMartificial}
\end{figure}

We thoroughly evaluate DCA-GMM on synthetic data generated with different variance level and noise level. The higher of these two levels, the harder the clustering task is. The results are reported in Figure \ref{fig:GMMartificial}, the detailed procedure generating data and evaluation metrics are given in the caption. On all metrics, DCA-GMM shows a phase transition property, i.e., the algorithm will overwhelmingly success below a curve of noise and variance level. This property verifies the robustness of DCA-GMM to data noise and variance within a cluster. In addition, when increasing the number of sub-problems (layers from bottom to top), the accuracy of DCA-GMM soon saturates on a value close to $1$, this indicates that a small number of sub-problems in DCA-GMM is sufficient to produce a promising clustering result, which is highly preferred in practice. Moreover, the time cost of DCA-GMM is significantly small and thus exhibits its competitive efficiency. Furthermore, error is more robust to data noise than accuracy, because most false anchors detected in the noise case are close to the true ones, which leads to small error.

\subsection{DCA for Gaussian Mixture Model on Image Dataset}. 

The experimental comparison results are shown in Figure \ref{fig:GMM}. DCA consistently outperforms other methods on accuracy on lots of datasets, and shows $20-2000$ times of acceleration in speed. By increasing the number of sub-problems, the accuracy of DCA improves. Note the pixels of face/handwritten digit/object images always exceed $1000$, and thus results in slow computation of pairwise distances required by other clustering methods. DCA exhibits the fastest speed because the number of sub-problems $s=\mathcal O(k\log k)$ does not depend on the feature dimension, and thus merely $171$ 2D random projections are sufficient for obtaining a promising clustering result. Spectral method performs poorer than DCA due to the large variance of sample moment. Because DCA uses the separability assumption as regularization in estimating the eigenspace of the moment, the variance is reduced.

\begin{figure}[htp]
\begin{center}
\subfigure{
 \includegraphics[width=1\linewidth]{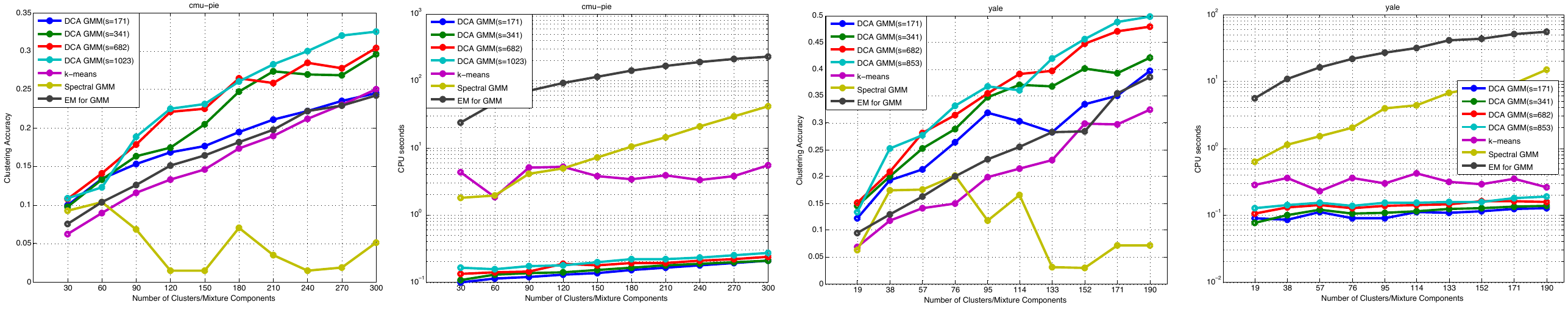}
 \label{fig:GMM1}
}
\subfigure{
 \includegraphics[width=1\linewidth]{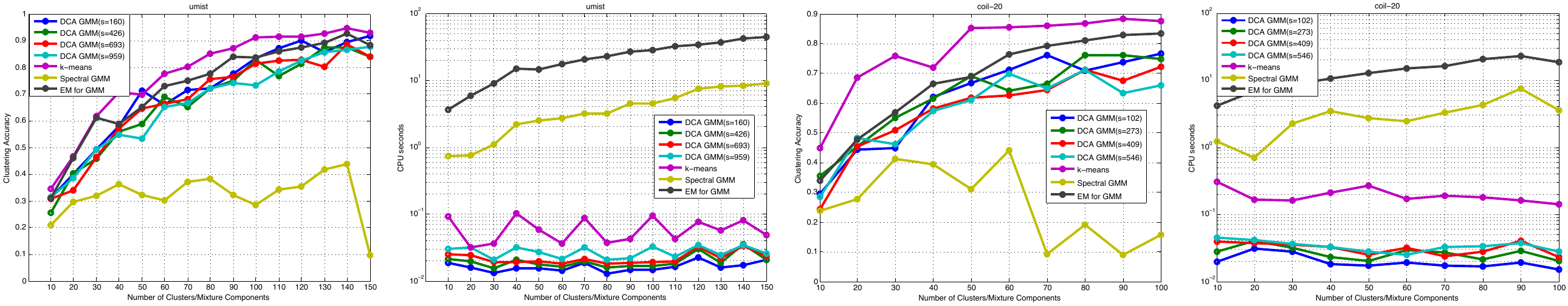}
 \label{fig:GMM2}
}
\subfigure{
 \includegraphics[width=1\linewidth]{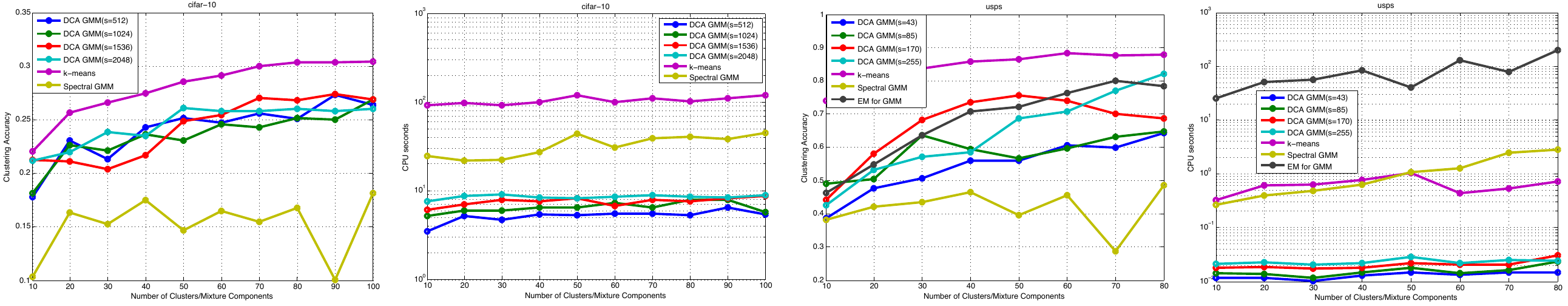}
 \label{fig:GMM3}
}
\end{center}
   \caption{Clustering accuracy (higher is better) and CPU seconds vs. Number of clusters for Gaussian mixture model on CMU-PIE, YALE, and UMIST human face datasets, UPSP handwritten digit dataset, CIFAR-10 image dataset, and COIL-20 object image dataset. We randomly split the raw pixel features into three groups, each associates to a view in our multi-view model. Baselines: K-means \cite{kmeans}, EM algorithm, spectral method.}
\label{fig:GMM}
\end{figure}

\subsection{DCA for Hidden Markov Model on Stock Price and Motion Capture Data}

The experimental comparison results for stock price modeling and motion segmentation are shown in Figure \ref{fig:HMMstock} and Figure \ref{fig:HMMmocap}, respectively. In the former one, DCA always achieves slightly lower but comparable likelihood compared to Baum-Welch (EM) method, while spectral method performs worse and unstably. DCA shows significant speed advantage compared to other methods, and thus is more preferable in practice. 

\begin{figure}[htp]
\begin{center}
\subfigure{
 \includegraphics[width=1\linewidth]{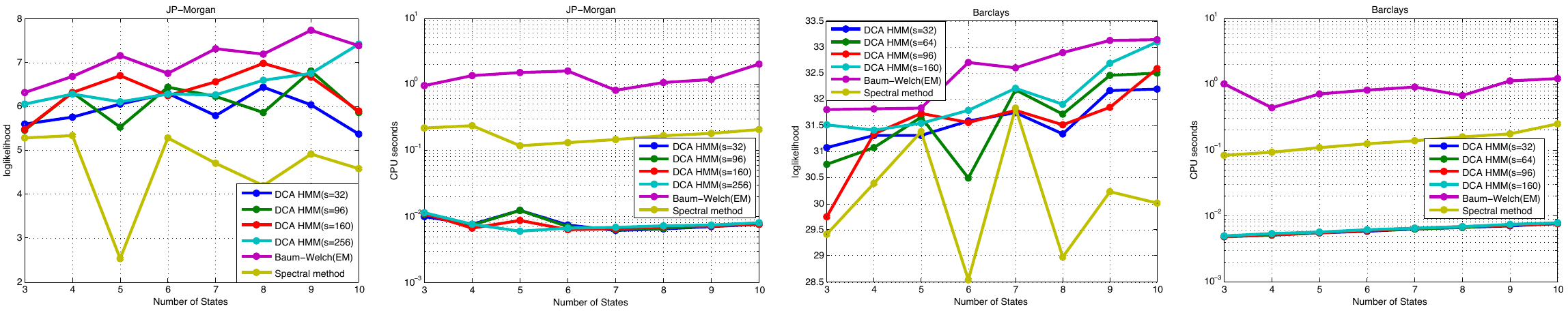}
 \label{fig:HMM1}
}
\subfigure{
 \includegraphics[width=1\linewidth]{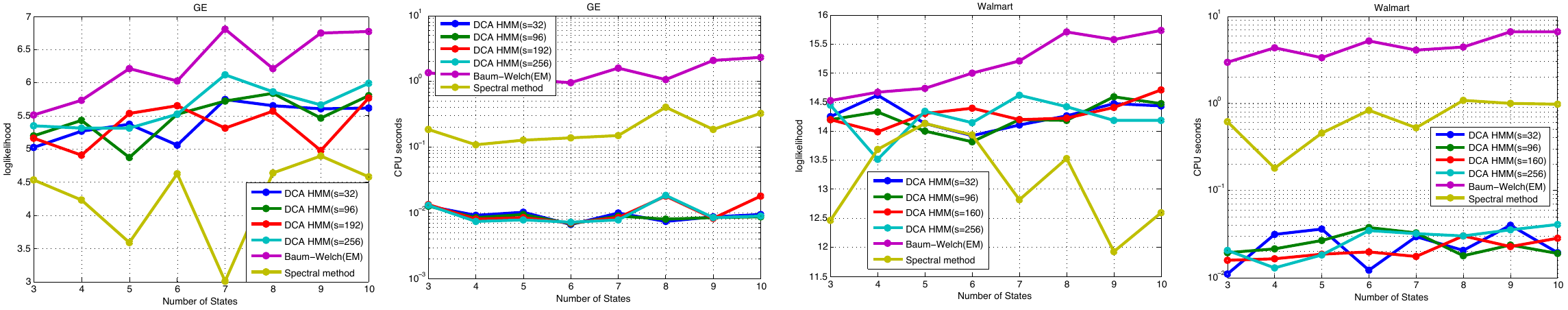}
 \label{fig:HMM2}
}
\end{center}
  \caption{Likelihood (higher is better) and CPU seconds vs. Number of states for using HMM to model stock price of $6$ companies from 01/01/1995-05/18/2014 collected by Yahoo Finance. Since no ground truth label is given, we can only measure the likelihood on training data. Baselines: Baum-Welch (EM) algorithm, spectral method.}
\label{fig:HMMstock}
\end{figure}

In the latter one, we evaluate the likelihood and prediction accuracy on both the training and the test set, so the regularization caused by separability assumption leads to the highest test accuracy and fastest speed of DCA. Note that the time cost of Baum-Welch method does not keeping increasing with the number of training samples in a constant speed. This is because the method uses an adaptive stop criterion, i.e., stop the optimization when the likelihood on the training data increases too slow. 

Since we cannot randomly select observations in a sequence for training due to the sequential property of the data, and due to the randomness in DCA, It is normal that the accuracy curve is not smooth. However, it is not hard to see that DCA usually achieves the highest accuracy given different number of training observations.

\begin{figure}[htp]
\begin{center}
\subfigure{
 \includegraphics[width=0.8\linewidth]{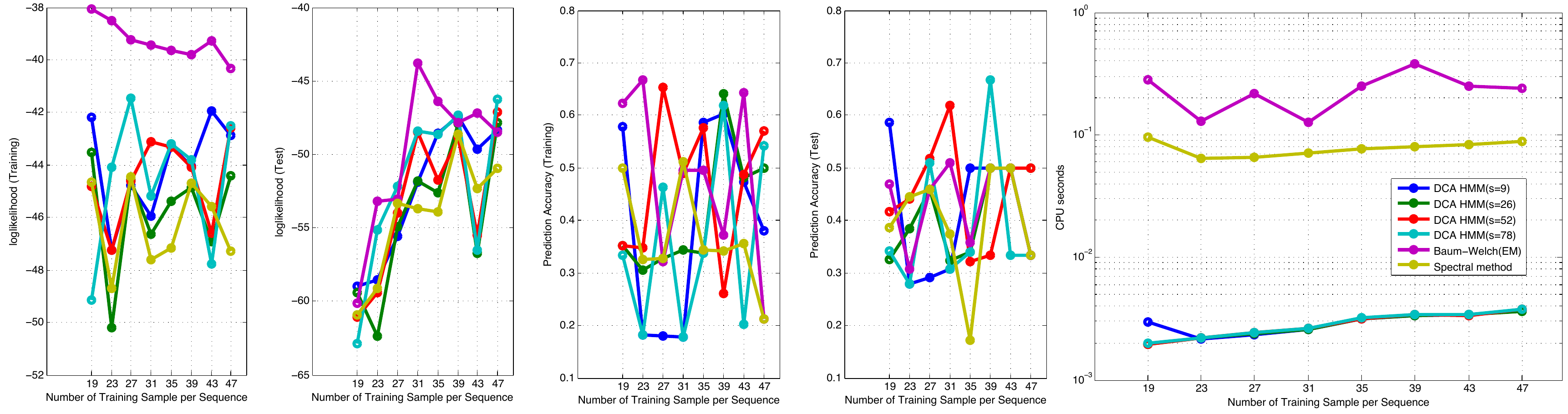}
 \label{fig:HMM3}
}
\subfigure{
 \includegraphics[width=0.8\linewidth]{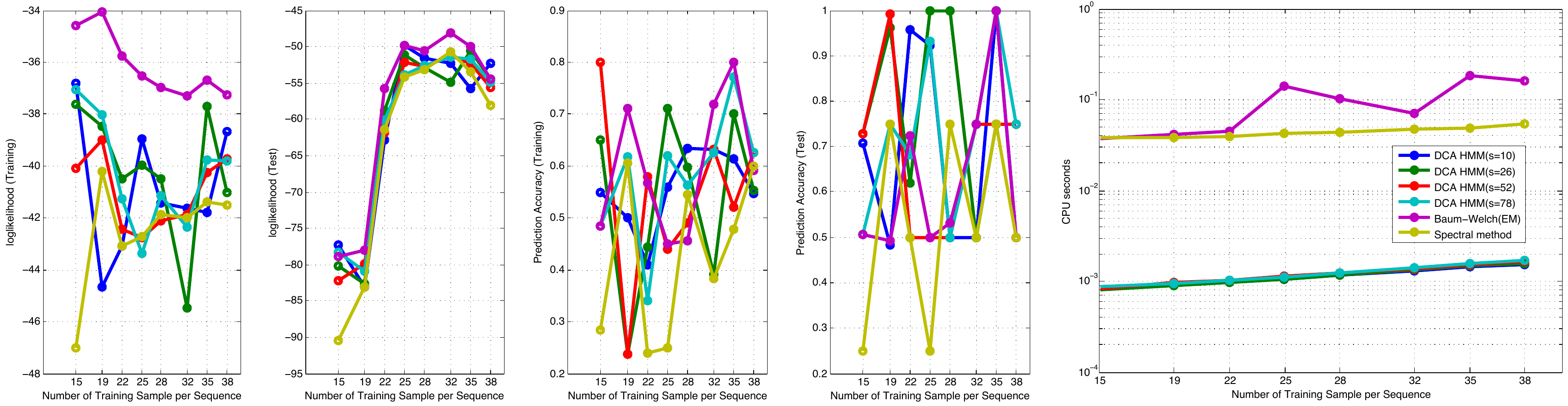}
 \label{fig:HMM4}
}
\subfigure{
 \includegraphics[width=0.8\linewidth]{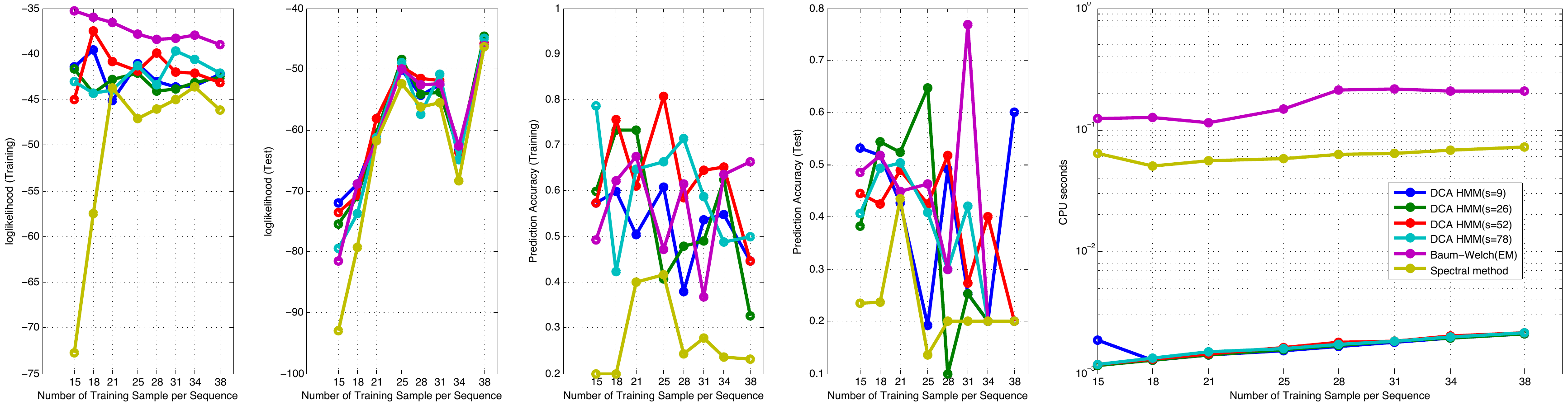}
 \label{fig:HMM5}
}
\subfigure{
 \includegraphics[width=0.8\linewidth]{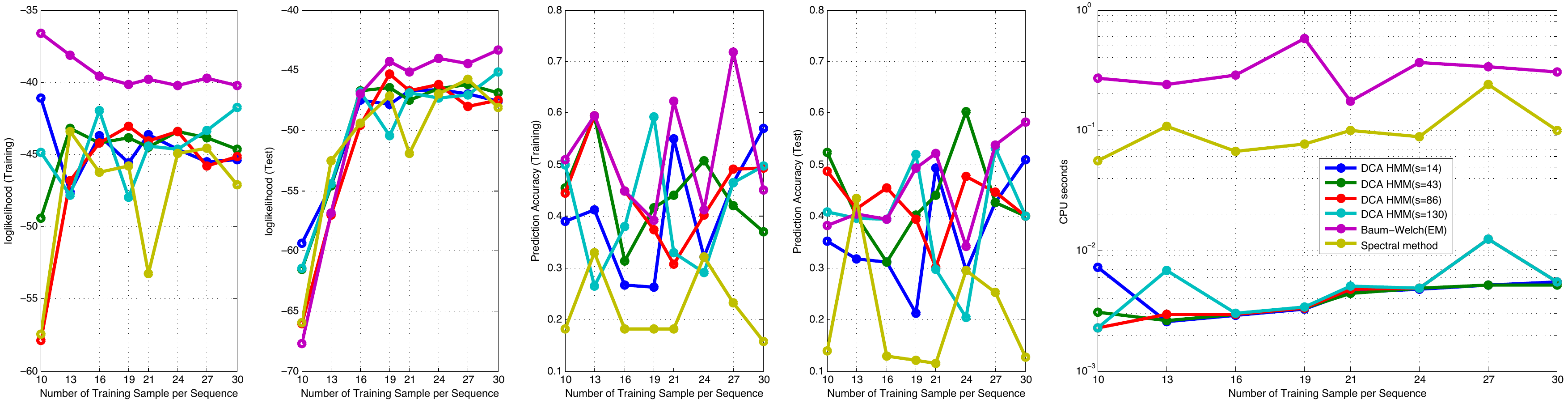}
 \label{fig:HMM6}
}
\subfigure{
 \includegraphics[width=0.8\linewidth]{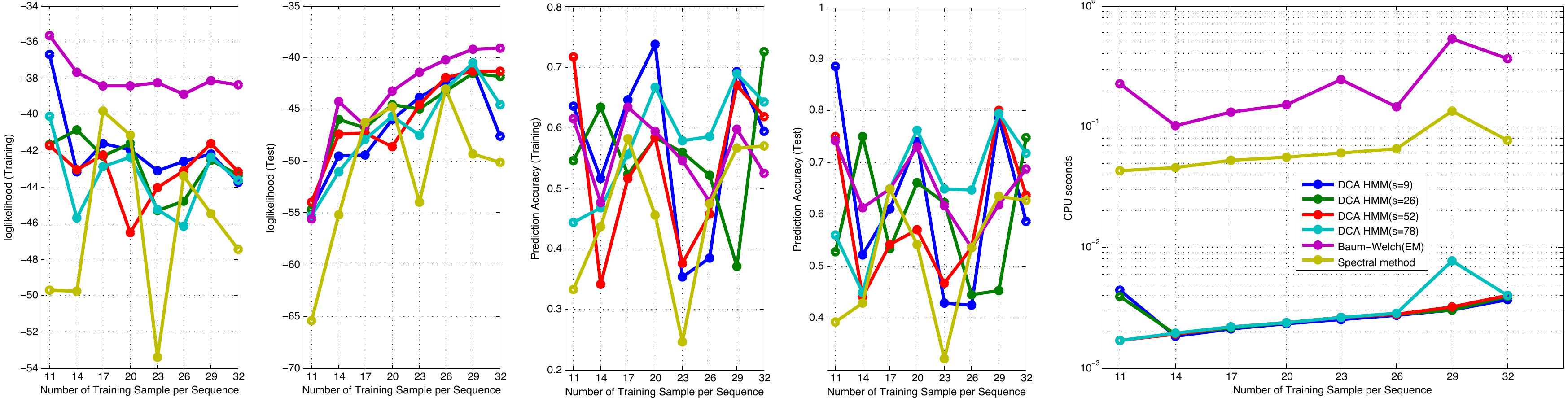}
 \label{fig:HMM7}
}
\subfigure{
 \includegraphics[width=0.8\linewidth]{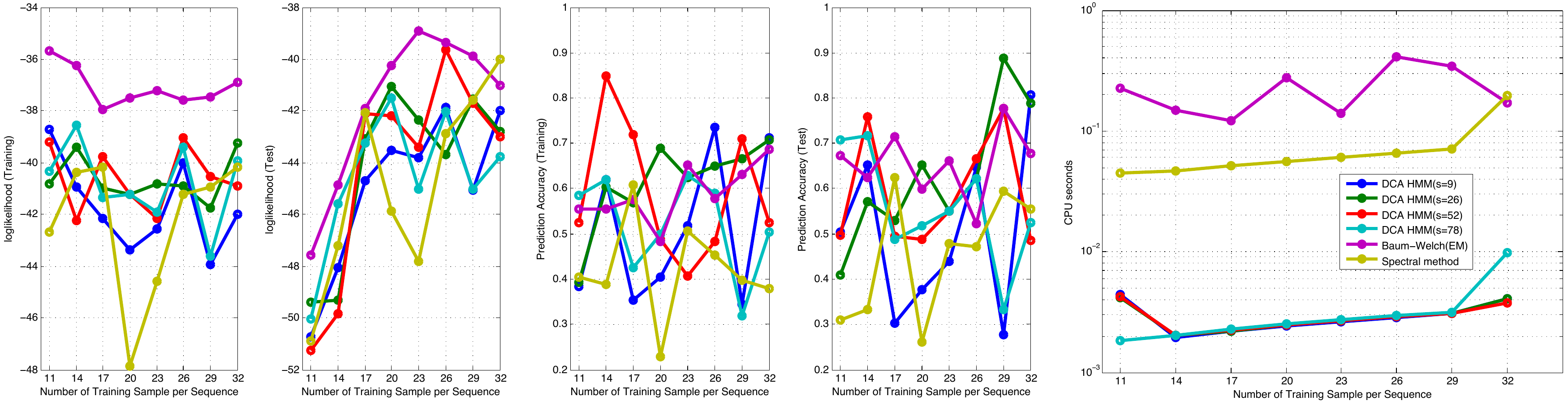}
 \label{fig:HMM8}
}
\end{center}
  \caption{Motion prediction training/test likelihood (higher is better), accuracy (higher is better) and CPU seconds for $6$ motion capture sequences from CMU-mocap dataset, under different number of training observations. The motion for each frame is manually labeled by the authors of \cite{BPHMM}. The total number of different motions in the $6$ sequences are $10,6,10,11,6,12$ respectively. Baselines: Baum-Welch (EM) algorithm, spectral method.}
\label{fig:HMMmocap}
\end{figure}

\subsection{DCA for Latent Dirichlet Allocation on Text Dataset}

The experimental comparison results for topic modeling are shown in Figure \ref{fig:LDA}. Compared to both traditional EM and the sampling method, DCA not only achieves both the smallest perplexity (highest likelihood) on the test set and the highest speed, but also the most stable performance when increasing the number of topics. In addition, the ``anchor word'' achieved by DCA provides more interpretable topics than other methods.

\begin{figure}[htb]
\begin{center}
 \includegraphics[width=1\linewidth]{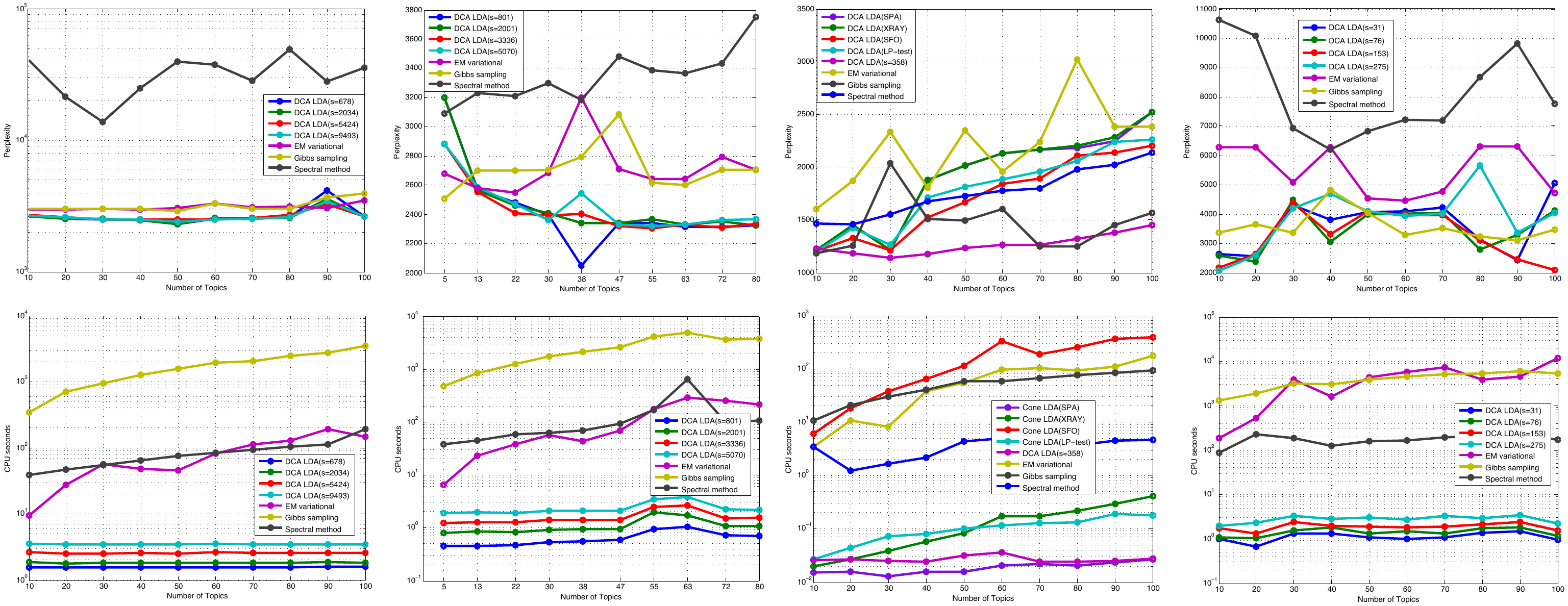}
\end{center}
   \caption{Perplexity (smaller is better) on test set and CPU seconds vs. Number of topics for LDA on NIPS, NIPS1-17, physical review and Grolier Dataset, we randomly selected $70\%$ documents for training and the rest $30\%$ is used for test. Baselines: EM algorithm for variational method, Gibbs sampling~\cite{fastLDA}, spectral method.}
\label{fig:LDA}
\end{figure}

\subsection{DCA for Subspace Clustering on Synthetic Dataset}

\begin{figure}[htb]
\begin{center}
 \includegraphics[width=1\linewidth]{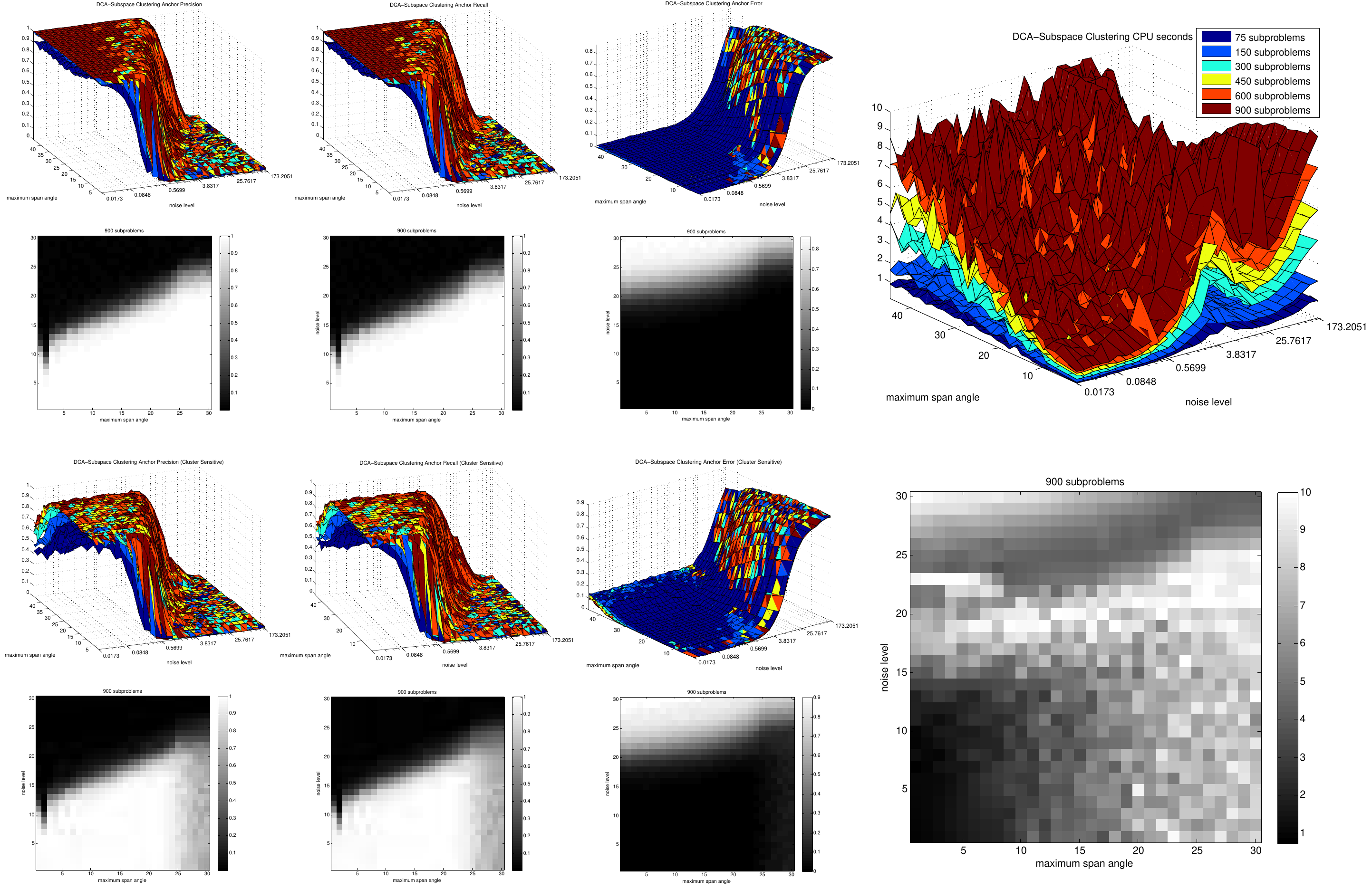}
\end{center}
   \caption{DCA-subspace clustering on synthetic data. On a $30\times 30$ grid of different noise level and maximum span angle between anchors defining different cones, for each pair, we randomly generate data of $k=4$ clusters (cones) with $500$ samples of dimension $300$ and $10$ extreme rays (anchors) per cluster. In particular, the conical combination coefficients of each point (corresponding to the $10$ anchors) in each cluster are drawn from a uniform distribution between $0$ and $1$, and then Gaussian noises of magnitude equal to the noise level are added to the points. We run DCA-subspace clustering using different number of sub-problems, and report their performance by precision, recall and error of the total $40$ anchors (ignore the wrong assignment of true anchor to a wrong cluster), and those metrics sensitive to clusters. The CPU seconds are reported too. Each point on each 3D plot is a result of averaging $10$ random trials in the same setting, and its height is the value of metric. Each 3D plot includes layers of surfaces associated with different number of sub-problems, we also report the top layer as a 2D plot below each 3D plot.}
\label{fig:SCartificial}
\end{figure}

We thoroughly evaluate DCA-subspace clustering on synthetic data generated with different noise level and maximum span angle, which is the maximum angle between two anchors of two conical hulls. Hence, large maximum span angle leads to a hard clustering problem. The results are reported in Figure \ref{fig:SCartificial}, the detailed procedure generating data and evaluation metrics are given in the caption. On all metrics, DCA-subspace clustering shows a phase transition property, i.e., the algorithm will overwhelmingly success below a curve of noise level and maximum span angle. This property verifies the robustness of DCA-subspace clustering to data noise and overlapping between cluster. In addition, when increasing the number of sub-problems (layers from bottom to top), the precision/recall of DCA-subspace clustering soon saturates on a value close to $1$, this indicates that a small number of sub-problems is sufficient to produce a promising clustering result, which is highly preferred in practice. Moreover, the time cost of DCA-subspace clustering is significantly small and thus exhibits its competitive efficiency. Furthermore, error is more robust to data noise than precision/recall, because most false anchors detected in the noise case are close to the true ones, which leads to small error.

It is worth noting that the same metric for all anchors and for cluster sensitive case shows different behaviors in the region of ``large maximum span angle, low noise level''. In particular, the precision/recall in cluster sensitive case decreases in this region, because although small noise level improves the probability of successfully identifying the anchors, the overlapping between cones caused by the large maximum span angle will lead to wrong assignment of anchors to clusters.

\subsection{DCA for Subspace Clustering on Image and Motion Capture Dataset}

The experimental comparison results for subspace clustering on object image dataset COIL-100 are shown in Figure \ref{fig:SC}. DCA provides a much more practical algorithm in speed that can achieve comparable mutual information but more than $1000$ times speedup than the state-of-the-art SC algorithms~\cite{SCC,SSC,LRR,RSC}. 

\begin{figure}[htb]
\begin{center}
 \includegraphics[width=1\linewidth]{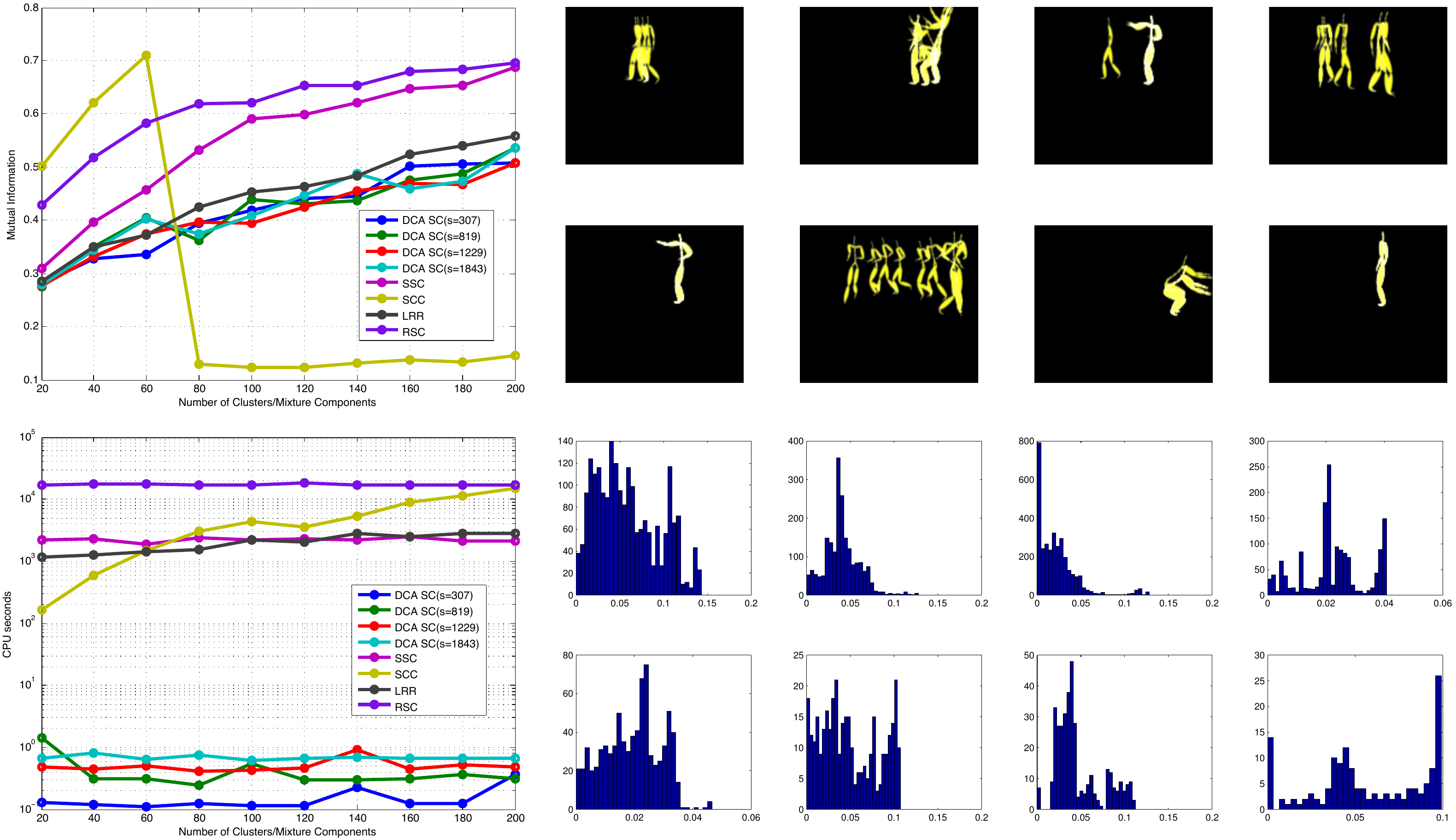}
\end{center}
   \caption{LEFT: Mutual Information (higher is better) and CPU seconds vs. Number of clusters for subspace clustering on COIL-100 object image Dataset. RIGHT: anchor frames for the $8$ clusters detected by DCA-subspace clustering, and the $\ell_2$ relative reconstruction error of frames in each cluster by using the detected anchors for the cluster. The data is sequence 2 of subject 86 from CMU-mocap dataset, with $62$ features collected from sensors on human body, for $>10000$ frames. We set the number of all anchors to be $50$ and the number of clusters to be $8$.}
\label{fig:SC}
\end{figure}

We also apply DCA-subspace clustering to a sequence of motion capture data that cannot be analyzed by existing subspace clustering methods due to their high computational complexity. DCA-subspace clustering aims to find several anchor frames for each cluster such that they can reconstruct most of the frames in the same cluster as their conical combinations. According to the results exhibited in Figure \ref{fig:SC}, the anchor frames in each of the $8$ detected clusters summarize one kind of motion on critical positions. So DCA provides a more interpretable subspace clustering results than other methods which usually define each cluster by several artificial bases. In addition, the reconstruction error for each cluster is small, and indicates that the selected anchor frames in each cluster are expressive and successfully summarize the associated motion. It also is worth noting that DCA-subspace clustering only costs $2.7$s to obtain the results, which is much less than the time costs of most the state-of-the-art approaches.

\section{Conclusion}

In this paper, we propose a general scheme that can reduce the parameter learning for a broad class of models, e.g., matrix factorization and latent variable model, to a geometric problem that aims to find a limited number of extreme rays so called ``anchors'' of a conical hull from a finite set of real data points. Compared to EM and sampling, which are the dominating parameter learning methods nowadays, our approach avoids alternating updating between parameter and latent variables, does not require iterative procedures, and provide a global solution guarantee based on the identifiability of the anchor set. By generalizing the separability assumption $X=FX_A$ for unique NMF to a more adaptive case $X=FY_A$, we propose a general minimum conical hull problem to formulate the reduced problem, and give rigorous theoretical analysis to the identifiability and uniqueness of its solution, as well as its interesting connections to other problems such as submodular set cover problem. As examples, we give the details of how to reduce learning NMF, subspace clustering, GMM, HMM, Kalman filter and LDA to the general minimum conical hull problem.

We further show that a novel idea of divide-and-conquer randomization leads to a significant efficient algorithm scheme for general minimum conical hull problem. In this ``divide-and-conquer anchoring (DCA)'' scheme, the original anchor finding task is distributed to multiple same-type sub-problems, each of which aims to find anchors (of the minimal conical hull) on a low-dimensional hyperplane, where the random projections of all data points lie in. Different from other randomized algorithms, each sub-problem in DCA only guarantees to recover a subset of anchors. This weaker requirement allows us to project the data points into extremely low-D hyperplane. But due to randomness, combining the anchors found in all sub-problems gives an accurate estimate of the true anchor set \emph{w.h.p.}. Rigorous analysis shows that the number of sub-problem to achieve such probabilistic guarantee is merely $\mathcal O(k\log k)$, where $k$ is the number of true anchors. Since we can apply any solver to the sub-problem, DCA provides a unified scheme solving general minimum conical hull problem. In addition, since most existing solvers have super-linear time complexity w.r.t. the data dimension, the algorithm generated by DCA invoking a solver is much faster than the solver itself. Furthermore, we show that DCA can be also used as a subroutine in other methods to provide an usually faster test checking if a point is covered by a conical hull or not.

In the special case when the hyperplane in each sub-problem is a 2D plane, we develop an ultrafast solver that precisely identifies the two 2D anchors by only computing an array of cosine values and finding the max/min values in a sub-array. Compared to the solvers relying on iterative optimization or sampling, our solver is simpler in implementation and faster in speed. Therefore, plugging it into DCA scheme produces a significantly effective DCA algorithm, which is later applied to all examples of learning algorithm design for specific models.

Comprehensive experiments on rich datasets and thorough comparison to the state-of-the-art algorithms for different learning tasks promisingly justify the significant improvement in speed and robustness brought by our approach. In particular, DCA algorithms for different specific models usually show tens to thousands times acceleration, and better generalization performance on test sets. Moreover, the anchors selected from real data points often provide more interpretable models and convincing explanations, which are preferred in real applications. 

\subsection{Future Works and Discussions} 

Although we present both the general minimum conical hull problem formulation for general learning models, and the unified scheme of DCA for solving the problem with detailed examples on popular specific models, there are several interesting and important potential extended topics of our method.

\begin{itemize}
\item In order to break the linearity assumption $\mathbb E(x|h)=h^TO^T$ and generalize distribution $p(x|h)$ to even non-parametric forms, we can consider to embed the joint distribution of $\{x_i\}_{i\in[3]}$ into a \emph{reproducing kernel Hilbert space (RKHS)}. Hence we can instead assume  $\mathbb E(x|h)=f(h)$, where function $f(\cdot)$ is an point in RKHS. Accordingly, the moments of finite feature vectors in (\ref{equ: moments23}) becomes moment operators of feature functions (or infinite feature vectors). By using the kernel trick introduced by reproducing property, the learning problem can be solved in $n$-D space in the same way as methods proposed in this paper. Note the same trick can be used to model and solve kernel matrix factorization in functional space too.

\item In higher order graphical models (e.g., n-gram models, higher order MRF and CRF) engaging more complicated structures, we usually parameterize the models with higher order (conditional) moments of $p(x|h)$, rather than conditional mean. By using the trick replacing $x_i$ in (\ref{equ: moments23}) with ${\rm vec}(x_i\otimes^n)$, the vectorization of the $n^{th}$ tensor power of $x_i$, columns of $O$ stores the conditional moments and can be recovered by using our approach in this paper.

\item In hierarchical graphical model (e.g., Bayesian networks) where $x\sim p(x|h_1)$ and $h_t\sim p(h_t|h_{t-1})$, we can apply our method to each layer of the model in a bottom-up learning manner and learn the parameters for each layer sequentially by chain rule. A very related work~\cite{SpectralMM} has shown this possibility for linear Bayesian networks. However, the estimation bias can be propagated throughout the learning process and leads to poor estimation of the parameters in higher layers. The general separability assumption in our method is capable to provide effective regularizations reducing the increasing bias. 

\item The paper also provides an interesting potential solver to \emph{semi-definite programming (SDP)} when the matrix variable $X$ has a low-rank penalty/constraint, which is exactly the case for many popular machine learning models. The essential idea is to represent $X$ by a weighted sum of $k$ rank-one matrices~\cite{RCP}, each of which is generated by a real data point. In optimization, we can either randomly select multiple rank-one matrices according to some probability, or select them in a greedy fashion according to certain score. In most situations, the probability or score is proportional to the probability of being anchors, and hence can be quickly obtained by solving a general minimum conical hull problem by DCA. This optimization approach is able to produce a more interpretable low-rank solution with faster speed.
\end{itemize}

We believe that inspired by the new insights of this paper in both problem formulation and algorithm design, the learning process of various machine learning models can be largely simplified and significantly accelerated. In addition, the learning results can become more convincing and explainable even for users outside machine learning community.

}{See~\cite{nips2014supplemental} for a complete experimental section with results of DCA for NMF, SC, GMM, HMM, and LDA, and comparison to other methods on more synthetic and real datasets.}

\notarxiv{

\begin{figure}[htp]\vspace{-2mm}
\begin{center}
 \includegraphics[width=0.7\linewidth]{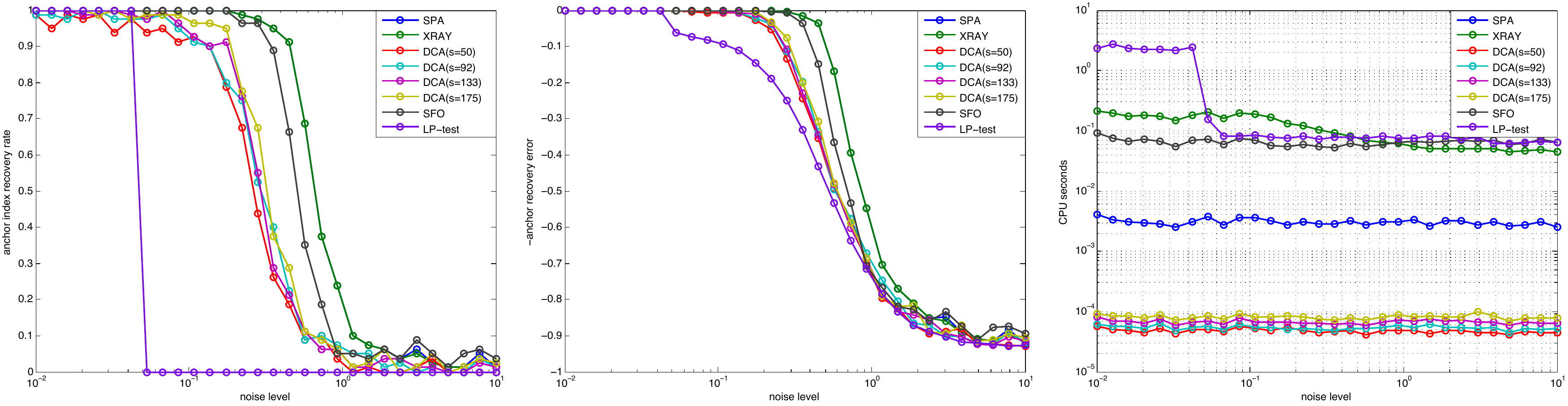}
\end{center}\vspace{-4mm}
   \caption{\scriptsize{Separable NMF on randomly generated $300\times 500$ matrix, each point on each curve is the result by averaging $10$ independent random trials. SFO-greedy algorithm for submodular set cover problem. LP-test is the backward removal algorithm from~\cite{AGKM}. LEFT: Accuracy of anchor detection (higher is better). Middle: Negative relative $\ell_2$ recovery error of anchors (higher is better). Right: CPU seconds.}}
\label{fig:NMFshort}\vspace{-2mm}
\end{figure}

\begin{figure}[htp]\vspace{-2mm}
 \begin{center}
  \includegraphics[width=1\linewidth]{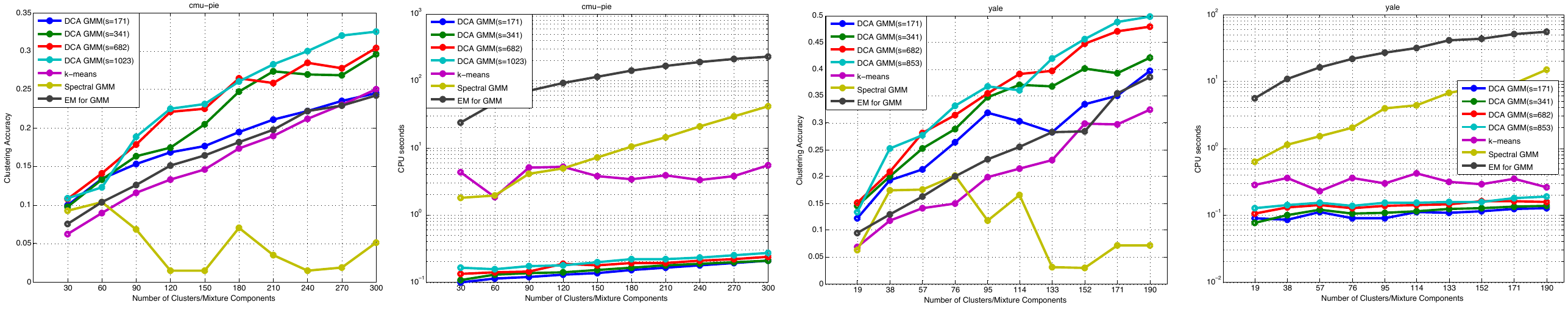}
 \end{center}\vspace{-4mm}
   \caption{\scriptsize{Clustering accuracy (higher is better) and CPU seconds vs. Number of clusters for Gaussian mixture model on CMU-PIE (left) and YALE (right) human face dataset. We randomly split the raw pixel features into 3 groups, each associates to a view in our multi-view model.}}
 \label{fig:GMMshort}
 \end{figure}\vspace{-2mm}
 
 \begin{figure}[htp]\vspace{-3mm}
 \begin{center}
  \includegraphics[width=1\linewidth]{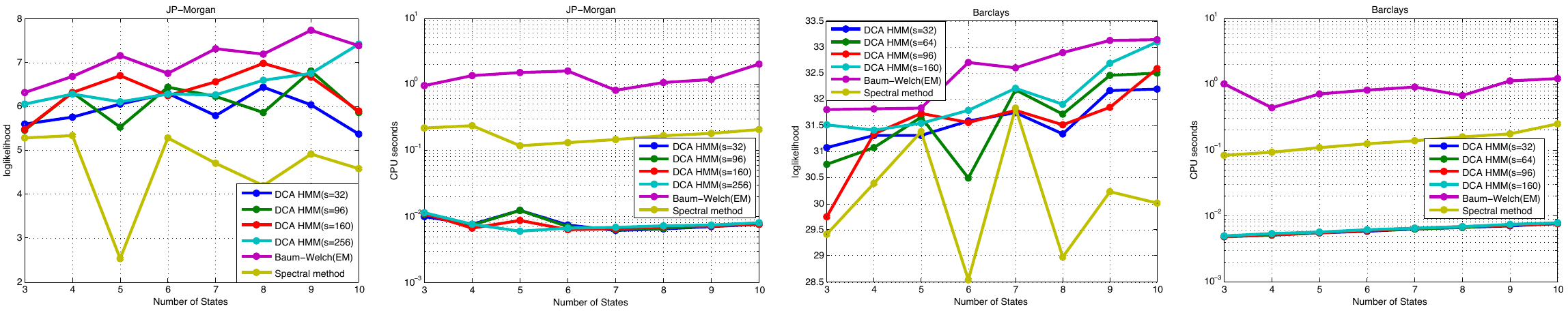}
 \end{center}\vspace{-2mm}
   \caption{\scriptsize{Likelihood (higher is better) and CPU seconds vs. Number of states for using HMM to model stock price of $2$ companies from 01/01/1995-05/18/2014 collected by Yahoo Finance. Since no ground truth label is given, we can only measure the likelihood on training data.}}
 \label{fig:HMMstockshort}
 \vspace{-7mm}
 \end{figure}\vspace{-2mm}

\textbf{DCA for Non-negative Matrix Factorization on Synthetic Data}. The experimental comparison results are shown in Figure \ref{fig:NMFshort}. Greedy algorithms SPA~\cite{SPA}, XRAY~\cite{XRAY} and SFO achieves the best accuracy and smallest recovery error when noise level is above $0.2$, but XRAY and SFO are the slowest two. SPA is slightly faster but still much slower than DCA. DCA with different number of sub-problems shows slightly less accuracy than greedy algorithms, but the difference is acceptable. Considering its significant acceleration, DCA offers an advantageous trade-off. LP-test~\cite{AGKM} has the exact solution guarantee, but it is not robust to noise, and too slow in speed. Therefore, DCA provides a much faster and more practical NMF algorithm with comparable performance to the best ones.

\textbf{DCA for Gaussian Mixture Model on CMU-PIE and YALE Face Dataset}. The experimental comparison results are shown in Figure \ref{fig:GMMshort}. DCA consistently outperforms other methods (k-means, EM, spectral method~\cite{SpectralLDA}) on accuracy, and shows $20$-$2000$ times acceleration in speed. By increasing the number of sub-problems, the accuracy of DCA improves. Note the pixels of face images always exceed $1000$, and thus results in slow computation of pairwise distances required by other clustering methods. DCA exhibits the fastest speed because the number of sub-problems $s=\mathcal O(k\log k)$ does not depend on the feature dimension, and thus merely $171$ 2D random projections are sufficient for obtaining a promising clustering result. The spectral method performs poorer than DCA due to the large variance of sample moment. Because DCA uses separability assumption as regularization in estimating the eigenspace of the moment, and thus effectively  reduces the variance.
 
 \begin{table}[htp]\scriptsize
 \vspace{-2mm}
 \caption{\scriptsize{Motion prediction accuracy (higher is better) of the test set for $6$ motion capture sequences from CMU-mocap dataset. The motion for each frame is manually labeled by the authors of \cite{BPHMM}. In the table, s13s29(39/63) means that we split sequence 29 of subject 13 into sub-sequences, each has $63$ frames, in which the first $39$ ones are for training and the rest are for test. Time is measured in ms.}}
  \vspace{-4mm}
 \begin{center}
 \begin{tabular}{l*{12}{c}}
 \hline
 Sequence &\multicolumn{2}{c}{s13s29(39/63)} &\multicolumn{2}{c}{s13s30(25/51)} &\multicolumn{2}{c}{s13s31(25/50)} & \multicolumn{2}{c}{s14s06(24/40)} &\multicolumn{2}{c}{s14s14(29/43)} &\multicolumn{2}{c}{s14s20(29/43)}\\
 \hline
 Measure &Acc &Time &Acc &Time &Acc &Time &Acc &Time  &Acc &Time &Acc &Time\\
 \hline\hline
 Baum-Welch (EM) &0.50 &383 &0.50 &140 &0.46 &148 &0.34 &368 &0.62 &529 &0.77 &345	\\
 Spectral Method &0.20 &80 &0.25 &43 &0.13 &58 &0.29 &66 &0.63 &134 &0.59 &70	\\
 DCA-HMM (s=9)  &0.33 &3.3 &0.92 &1 &0.19 &1.5 &0.29 &4.8 &0.79 &3 &0.28 &3	\\
 DCA-HMM (s=26)  &0.50 &3.3 &\bf{1.00} &1 &\bf{0.65} &1.6 &\bf{0.60} &4.8 &0.45 &3 &\bf{0.89} &3	\\
 DCA-HMM (s=52)  &0.50 &3.4 &0.50  &1.1 &0.43 &1.6 &0.48 &4.9 &\bf{0.80} &3.2 &0.78 &3.1	\\
 DCA-HMM (s=78)  &\bf{0.66} &3.4 &0.93 &1.1 &0.41 &1.6 &0.51 &4.9 &\bf{0.80} &6.7 &0.83 &3.2	\\
 \hline
 \end{tabular}\label{table:hmmmocap}
 \end{center}\vspace{-2mm}
 \vspace{-2mm}
 \end{table}
 \normalsize
 
 \begin{figure}[htp]
 \begin{center}
  \includegraphics[width=1\linewidth]{LDASC.pdf}
 \end{center}\vspace{-4mm}
   \caption{\scriptsize{\textbf{LEFT:} Perplexity (smaller is better) on test set and CPU seconds vs. Number of topics for LDA on NIPS1-17 Dataset, we randomly selected $70\%$ documents for training and the rest $30\%$ is used for test. \textbf{RIGHT:} Mutual Information (higher is better) and CPU seconds vs. Number of clusters for subspace clustering on COIL-100 Dataset.}}
 \label{fig:LDASC}
 \end{figure}\vspace{-2mm}

\textbf{DCA for Hidden Markov Model on Stock Price and Motion Capture Data}. The experimental comparison results for stock price modeling and motion segmentation are shown in Figure \ref{fig:HMMstockshort} and Table \ref{table:hmmmocap}, respectively. In the former one, DCA always achieves slightly lower but comparable likelihood compared to Baum-Welch (EM) method~\cite{HMM}, while the spectral method~\cite{SpectralMM} performs worse and unstably. DCA shows a significant speed advantage compared to others, and thus is more preferable in practice. In the latter one, we evaluate the prediction accuracy on the test set, so the regularization caused by separability assumption leads to the highest accuracy and fastest speed of DCA.

\textbf{DCA for Latent Dirichlet Allocation on NIPS1-17 Dataset}. The experimental comparison results for topic modeling are shown in Figure \ref{fig:LDASC}. Compared to both traditional EM and the Gibbs sampling~\cite{fastLDA}, DCA not only achieves both the smallest perplexity (highest likelihood) on the test set and the highest speed, but also the most stable performance when increasing the number of topics. In addition, the ``anchor word'' achieved by DCA provides more interpretable topics than other methods.

\textbf{DCA for Subspace Clustering on COIL-100 Dataset}. The experimental comparison results for subspace clustering are shown in Figure \ref{fig:LDASC}. DCA provides a much more practical algorithm that can achieve comparable mutual information but more than $1000$ times speedup over the state-of-the-art SC algorithms such as SCC~\cite{SCC}, SSC~\cite{SSC}, LRR~\cite{LRR}, and RSC~\cite{RSC}. 

}

\small
\bibliographystyle{plain}
\bibliography{nips2014}

\end{document}